\documentclass[lettersize,journal]{IEEEtran}
\usepackage{amsmath,amsfonts}
\usepackage{algorithmic}
\usepackage{algorithm}
\usepackage{array}
\usepackage[caption=false,font=normalsize,labelfont=sf,textfont=sf]{subfig}
\usepackage{textcomp}
\usepackage{stfloats}
\usepackage{url}
\usepackage{graphicx}
\usepackage{cite}
\usepackage{titlesec}
\usepackage{subcaption}
\usepackage{multirow}
\usepackage{amssymb} % For \checkmark
\usepackage{xcolor} % xcolor
\usepackage[final,markup=underlined,nocomments]{changes} % draft or final
\usepackage{verbatim}

\usepackage{bm}
\usepackage{booktabs} % for \toprule, \midrule, \bottomrule

\usepackage[utf8]{inputenc}
\usepackage[T1]{fontenc}
\usepackage{makecell}
\usepackage{subfig}
\usepackage{adjustbox}

\renewcommand{\thesection}{\arabic{section}}
\renewcommand{\thesubsection}{\thesection.\arabic{subsection}}
\renewcommand{\thesubsubsection}{\thesubsection.\arabic{subsubsection}}

\titleformat{\section}[block]{\bfseries\Large}{\thesection.}{1em}{}
\titleformat{\subsection}[block]{\bfseries\large}{\thesubsection.}{1em}{}
\titleformat{\subsubsection}[block]{\bfseries\normalsize\itshape}{\thesubsubsection.}{1em}{}

\titlespacing*{\subsubsection}{0pt}{3.25ex plus 1ex minus .2ex}{0.5em}

\title{FilletRec: A Lightweight Graph Neural Network with Intrinsic Features for Automated Fillet Recognition}

\author{Jiali~Gao\textsuperscript{\dag}, Taoran~Liu\textsuperscript{\dag}, 
Hongfei~Ye\textsuperscript{*}
Jianjun~Chen\textsuperscript{*}

    \thanks{Jiali Gao is with the College of Computer Science and Technology, Zhejiang University and also with the State Key Lab of CAD\&CG, Zhejiang University, Hangzhou, 310027, China. E-mail: 22421004@zju.edu.cn. Taoran Liu, Hongfei Ye and Jianjun Chen are with the School of Aeronautics and Astronautics, Zhejiang University and also with the State Key Lab of CAD\&CG, Zhejiang University, Hangzhou, 310027, China. E-mail:\{taoranliu, hfye, chenjj\}@zju.edu.cn.} % 作者单位和邮箱
    \thanks{*Co-corresponding authors: Hongfei Ye, Jianjun Chen.} % 通讯作者说明
    \thanks{\textsuperscript{\dag}Co-first authors: Jiali Gao and Taoran Liu contributed equally to this work.} % 共一说明
}

\begin{document}

\maketitle % 生成标题

% =================== 摘要与关键词 ===================
\begin{abstract}
Automated recognition and simplification of fillet features in CAD models is critical for CAE analysis, yet it remains an open challenge. Traditional rule-based methods lack robustness, while existing deep learning models suffer from poor generalization and low accuracy on complex fillets due to their generic design and inadequate training data. To address these issues, this paper proposes an end-to-end, data-driven framework specifically for fillet features. We first construct and release a large-scale, diverse benchmark dataset for fillet recognition to address the inadequacy of existing data. Based on it, we propose FilletRec, a lightweight graph neural network. The core innovation of this network is its use of pose-invariant intrinsic geometric features, such as curvature, enabling it to learn more fundamental geometric patterns and thereby achieve high-precision recognition of complex geometric topologies. Experiments show that FilletRec surpasses state-of-the-art methods in both accuracy and generalization, while using only 0.2\%-5.4\% of the parameters of baseline models, demonstrating high model efficiency. Finally, the framework completes the automated workflow from recognition to simplification by integrating an effective geometric simplification algorithm.

\vspace{1em} % 在摘要和关键词之间增加一点垂直间距

\noindent % 防止段首缩进

\end{abstract}

\section{Introduction}

\begin{figure}[htbp]
    \centering
    \includegraphics[width=0.48\textwidth]{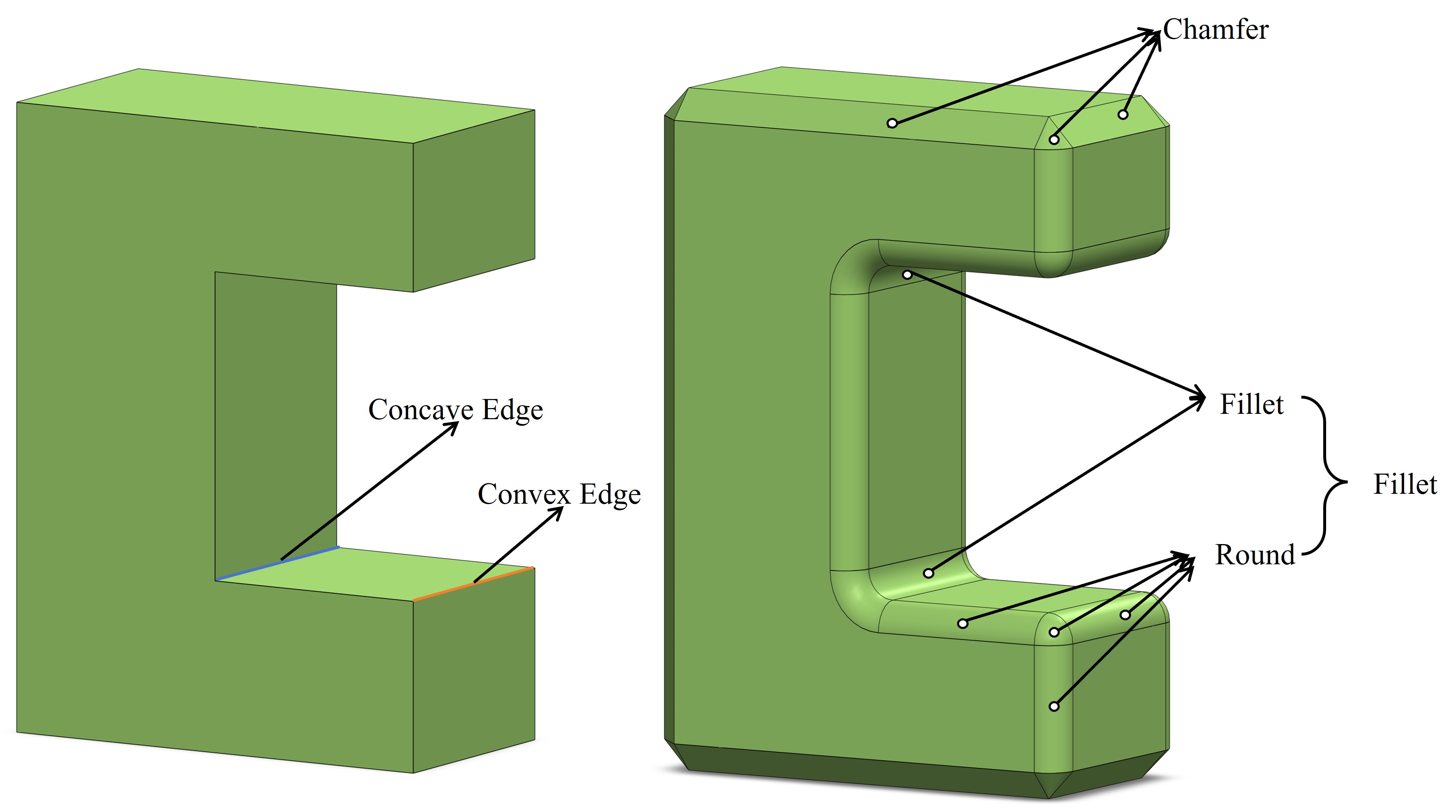} 
    \caption{Illustration of common edge features in a CAD model. The model on the left shows sharp concave and convex edges. The model on the right shows the same part after the application of fillets (on the concave edge), rounds (on the convex edges), and chamfers.}
    \label{fig:fillet_round}
\end{figure}

Fillets and rounds are geometric elements widely found in Computer-Aided Design (CAD) models to eliminate stress concentrations, satisfy manufacturing constraints, or meet aesthetic requirements \cite{pilkey2020peterson,enab2014stress,zhu2002brep}. As illustrated in Figure \ref{fig:fillet_round}, fillets are typically applied to concave edges, while rounds are applied to convex edges, though they are often collectively referred to as filleting features. While crucial in the design phase, they often become a significant bottleneck during Computer-Aided Engineering (CAE) analysis, especially for finite element meshing. Small-sized or complex fillets necessitate the generation of locally dense meshes, which not only significantly increases computational costs but can also lead to poor mesh quality, thereby compromising the accuracy and convergence of simulation analyses \cite{slyadnev2020simplification, lai2016small}. Therefore, in the CAE preprocessing workflow, the automatic recognition and simplification of these fillets is a critical step toward achieving an efficient and integrated design-to-analysis process.

\begin{figure}[htbp]
    \centering
    \includegraphics[width=0.48\textwidth]{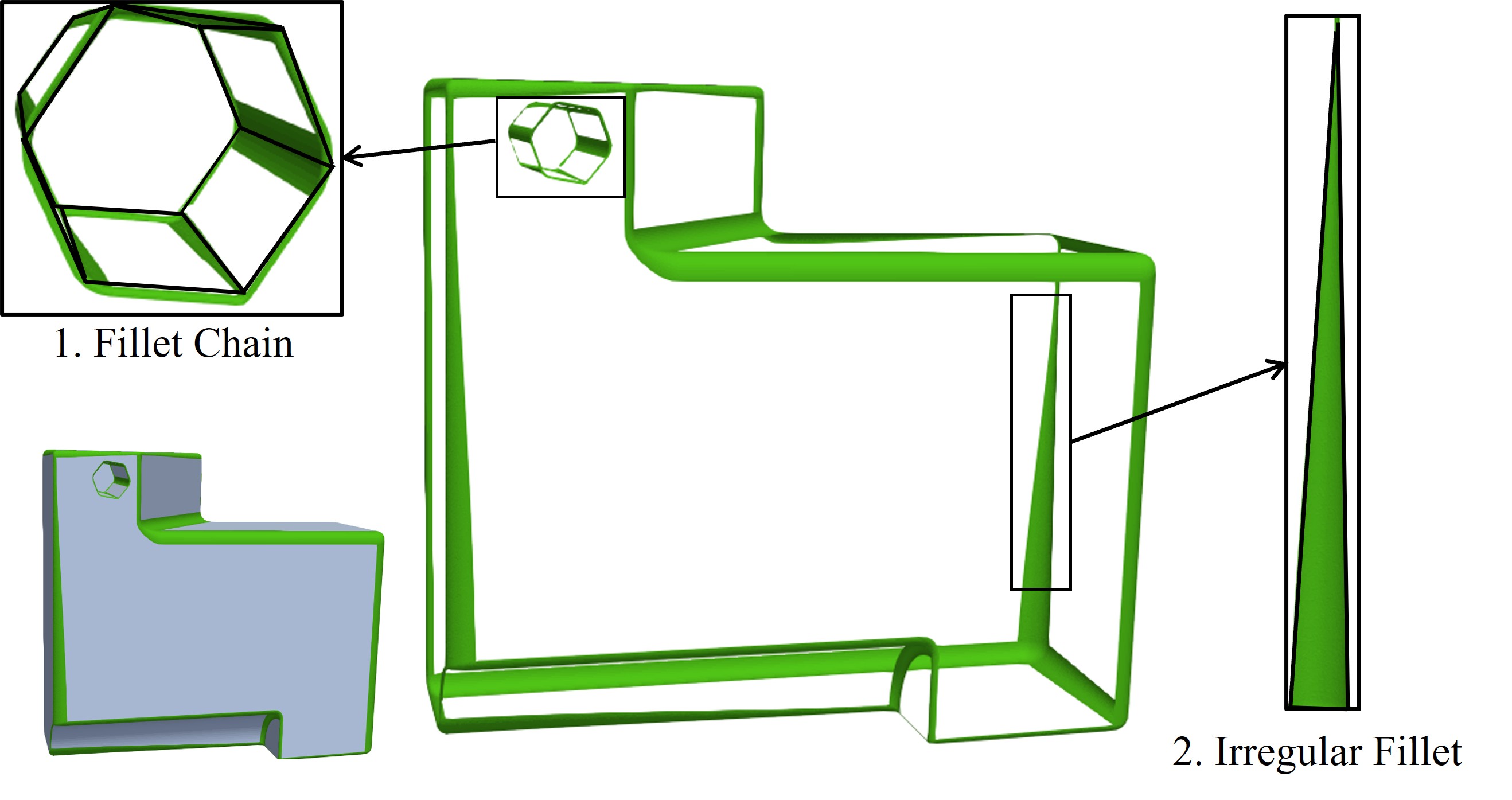} 
    \caption{Examples of fillet features.}
    \label{fig:fillet_sample}
\end{figure}

However, despite decades of research, the automated and robust recognition of fillet features remains an open challenge. Traditional methods are inherently rule-based, attempting to define and identify fillets through hand-crafted geometric property thresholds, topological pattern matching \cite{lai2016small,sunil2010approach}, or sequences of geometric operations \cite{zhu2002brep}. When faced with the complex situations prevalent in industrial models such as fillet chains  (as shown in Figure~\ref{fig:fillet_sample}) formed by the intersection of multiple fillets \cite{slyadnev2020simplification}, the robustness and generalization ability of these methods decline sharply.

In recent years, the rise of geometric deep learning \cite{bronstein2017geometric} has brought a paradigm shift to CAD feature recognition, achieving notable success in identifying manufacturing features \cite{jayaraman2021uvnet,colligan2022hierarchical, wu2024aagnet}. However, these advanced methods are designed to handle multiple feature types in a unified classification framework, and as a result, the unique properties of fillets as transitional features are often overlooked. The insufficient coverage and diversity of fillet samples in existing public datasets further exacerbates this challenge. Consequently, when these methods designed for general-purpose features are directly applied to fillet recognition tasks, their generalization ability is often inadequate. This is a core bottleneck for current data-driven methods in this specific application.

To address the aforementioned limitations of both traditional methods and deep learning-based approaches in fillet recognition, this paper proposes a neural network specifically designed for fillet feature recognition. First, to tackle the fundamental issue of data scarcity, we construct a large-scale, diverse benchmark dataset for fillet features, which contains numerous complex fillet cases commonly found in industrial scenarios. Building upon this dataset, we propose a lightweight graph neural network named FilletRec. The core innovation of this network lies in combining differential geometric properties with rigid transformation invariance as features with B-Rep topology information, enabling the network to learn more essential and robust geometric patterns. Furthermore, to achieve seamless integration between feature recognition results and geometric simplification, we design a robust Extend-Intersect-Clean simplification algorithm. This algorithm can directly process the complex fillet faces identified by FilletRec, automatically reconstruct sharp geometric edges, and ultimately generate simplified models that meet CAE analysis requirements \cite{thakur2009survey}.

The main contributions of this paper can be summarized as follows:
\begin{itemize}
    \item A large-scale and diverse benchmark dataset for fillet recognition\footnote{\raggedright Dataset and code: \url{https://ltrbless.github.io/projects/FilletRec}}. We construct and publicly release a comprehensive dataset containing 4,486 CAD models, covering various complex fillet types with precise face-level annotations. This dataset addresses the critical bottleneck of insufficient fillet samples and oversimplified geometric structures in existing datasets, providing the research community with a more challenging evaluation benchmark that better reflects industrial application scenarios.

    \item A lightweight graph neural network, FilletRec, based on intrinsic geometry. We propose a network designed for fillet recognition, centered on a feature representation that is invariant to rigid transformations. By combining differential geometric quantities like curvature with B-Rep topology, our model learns more fundamental geometric patterns, thereby achieving high-precision recognition on complex CAD models.

    \item An end-to-end automated defeaturing framework from recognition to simplification. We integrate the high-precision results from FilletRec with a robust Extend-Intersect-Clean geometric algorithm to build a complete automated workflow. This framework can directly process complex fillets identified by the FilletRec and generate simplified geometric models suitable for downstream CAE analysis, significantly improving the efficiency and reliability of engineering preprocessing.
\end{itemize}

\section{Related Work}

Research in feature recognition for CAD models has historically evolved along two distinct technical paths: the first relies on traditional methods based on precise geometric rules and topological reasoning, while the second involves emerging geometric deep learning approaches that automatically learn feature patterns through a data-driven approach. This section reviews the development, achievements, and inherent limitations of both approaches.

\subsection{Traditional Fillet Recognition Methods}
Traditional fillet recognition methods work directly on the geometric representation of a model and can be categorized into two main types based on the underlying data structure: those based on Boundary Representation (B-Rep) and those based on discrete meshes.

Methods based on B-Rep represent the most mature and deeply researched direction, with their advantage lying in the ability to leverage the precise topological relationships and analytical geometric information provided by B-Rep models. Early methods primarily relied on local geometric properties, such as detecting high curvature or changes in curvature continuity to screen for candidate fillet faces \cite{zhu2002brep, li2009automatic}. To enhance robustness, researchers further defined features as specific face-edge-vertex topological patterns and searched for them in the model's B-Rep graph using graph-matching algorithms \cite{Joshi1988GraphbasedHF}. For example, a simple fillet can be abstracted as a cylindrical or toroidal face tangent to two main faces. However, the core challenge of this paradigm is that the topological patterns of fillets in industrial models are extremely diverse, especially in complex "fillet chains" or "vertex fillet" formed by the intersection of multiple fillets. A predefined, finite library of topological rules struggles to cover all possible cases, leading to limited generality \cite{venkataraman2001blend, lai2016small}.

To address this challenge, subsequent research has focused on developing more generalized topological analysis techniques. For instance, Slyadnev and Turlapov \cite{slyadnev2020simplification} proposed an analysis method based on Euler operators to more robustly handle the suppression of complex fillet chains. For models containing free-form surfaces, Chen et al. \cite{chen2011fillet} utilized the concept of "spline chains" to identify and process more complex transitional features. More recently, Song et al. \cite{song2024recognition} combined "Spring Edge Groups" and the Attributed Adjacency Graph (AAG) to further improve recognition accuracy in complex scenarios.

Methods based on discrete meshes face greater challenges because the precise geometric and topological information is lost, forcing algorithms to rely solely on discrete vertices and faces for inference. These methods typically segment the mesh by estimating the curvature of discrete vertices or analyzing the variation in face normals to identify high-curvature fillet regions \cite{hase2021blend}. However, these strategies, based on local geometric estimation, are highly sensitive to mesh quality and noise. To move beyond reliance on local curvature, Jiang et al. \cite{jiang2025defillet} % Please check the year
proposed an innovative method that leverages the geometric property that vertices of the Voronoi diagram of surface sample points cluster densely along the trajectory of a rolling-ball's center to identify fillet regions, subsequently reconstructing sharp features through quadratic optimization. Although these methods have made progress in processing mesh data, they still face challenges in handling complex feature interactions, and their performance is highly dependent on the quality of the input mesh.

In summary, whether based on the precise topological analysis of B-Rep or the innovative geometric insights from meshes, traditional methods have demonstrated their effectiveness in handling specific types of fillets. However, their common bottleneck is the heavy reliance on hand-crafted rules, heuristics, or specific geometric invariants. This "prior knowledge-driven" paradigm struggles to generalize to the endless variety of complex geometric and topological configurations found in CAD models, making its robustness and level of automation insufficient to meet the demands of modern large-scale engineering analysis.

\subsection{Geometric Deep Learning for Feature Recognition}

In contrast to traditional methods, geometric deep learning aims to automatically learn feature representations from CAD models through a data-driven approach. The key challenge lies in designing a network architecture capable of effectively encoding the unique hybrid of topological and geometric information in B-Rep models. In recent years, Graph Neural Networks (GNNs) have proven to be a powerful tool for handling such data, as they can naturally map the topological relationships of a B-Rep model into a graph structure.

Early explorations framed feature recognition as a segmentation or classification problem on graph nodes (i.e., B-Rep faces).UV-Net \cite{jayaraman2021uvnet} is a representative work that samples a regular UV grid on each parametric surface, uses a CNN to extract geometric features, and simultaneously employs a GNN to propagate and aggregate information on the face adjacency graph, thus achieving an effective combination of geometry and topology. In pursuit of purer topological invariance, BRepNet \cite{lambourne2021brepnet} proposed a convolution method based on "topological walks," making it independent of coordinate information and inherently geometrically invariant. To more comprehensively fuse both types of information, Hierarchical CADNet \cite{colligan2022hierarchical} designed a hierarchical graph structure to learn features from both macroscopic topology and microscopic geometry. As research has progressed, the complexity of the tasks has also increased. AAGNet \cite{wu2024aagnet} implements multi-task learning by constructing a more information-rich attribute graph, enabling simultaneous semantic and instance segmentation.

However, despite their notable success, these methods share a common issue: they typically use the 3D coordinates of sampled surface points as the primary geometric input \cite{jayaraman2021uvnet,lambourne2021brepnet,colligan2022hierarchical,wu2024aagnet}. This extrinsic geometric representation is highly sensitive to the model's spatial pose (translation and rotation), which limits the model's generalization capabilities. To address this critical issue, we propose a recognition framework based on intrinsic geometry and graph neural networks, aiming to learn a more robust feature representation that is invariant to rigid transformations.

In summary, methods based on geometric deep learning demonstrate immense potential through end-to-end learning, especially in handling complicated and diverse data. However, the limitations of existing works in geometric representation leave a clear research gap for us to utilize intrinsic geometry to enhance model generalization and robustness.

\section{Method}

\begin{figure*}[htbp]
    \centering
    \includegraphics[width=0.9\textwidth]{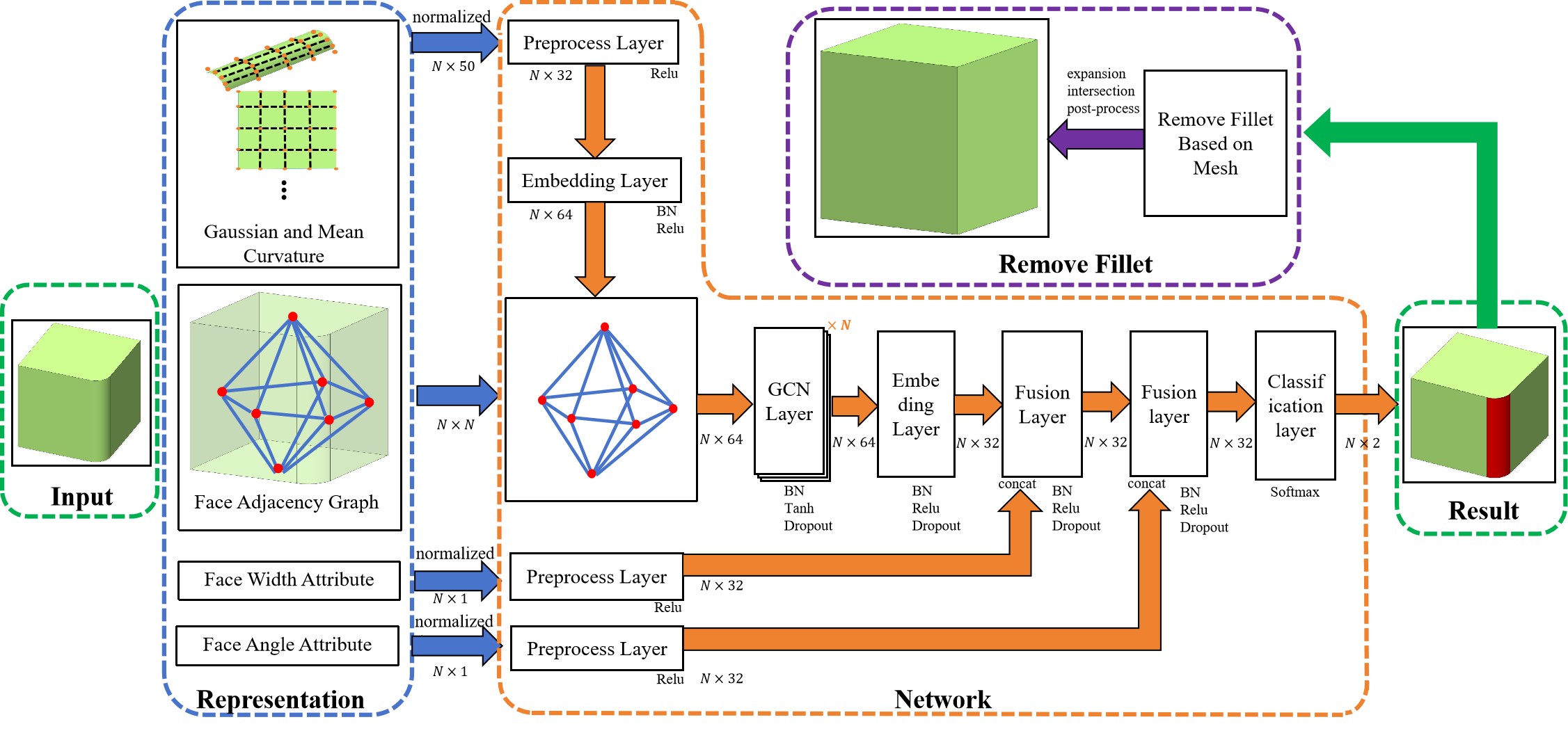} 
    \caption{The overall pipeline of our proposed method. Given a B-Rep model, we first construct a graph by mapping faces to nodes and adjacency to edges. We then extract intrinsic geometric features (curvature) and attribute features (face width, dihedral angle) for each node. Finally, these features are fed into our FilletRec network for binary classification to identify fillet faces.}
    \label{fig:overview}
\end{figure*}

Figure~\ref{fig:overview} illustrates the overall pipeline of our method. Given a B-Rep model composed of NURBS surfaces as input, our objective is to perform binary classification on each NURBS surface to determine whether it represents a fillet feature. This section elaborates on the construction of input representations (Section \ref{sec:input_representation}) and the design details of the network architecture (Section \ref{sec:network_architecture}).

\subsection{Input Representation}

\label{sec:input_representation}

Boundary Representation (B-Rep) is a widely adopted geometric modeling method in CAD systems~\cite{ZHU2002109}, which completely describes three-dimensional solids through faces, edges, vertices, and their topological relationships. To enable deep learning models to process B-Rep data, we convert it into a graph structure representation: mapping surfaces to graph nodes, face adjacency relationships to graph edges, and extracting multimodal feature vectors for each node.

Formally, given a B-Rep model $\mathcal{M}$, we construct an undirected graph $G = (V, E, X)$, where $V = \{\mathcal{S}_1, \mathcal{S}_2, \ldots, \mathcal{S}_n\}$ is the set of surfaces, $E \subseteq V \times V$ is the edge set encoding face adjacency relationships, and $X \in \mathbb{R}^{|V| \times d}$ is the node feature matrix. Node features consist of two components:
\begin{enumerate}
    \item \textbf{Geometric features}: characterizing the intrinsic shape properties of surfaces;
    \item \textbf{Attribute features}: capturing macroscopic information with engineering semantics.
\end{enumerate}

The extraction methods for each type of feature are introduced below.

\subsubsection{Topological Features}

The types of geometric features are closely related to the topological relationships between surfaces. For example, fillet features consist of transition surfaces connecting two primary faces, while hole features have inner surfaces forming closed topological loops. Therefore, face adjacency relationships serve as crucial prior information for feature recognition.

We model face adjacency relationships as an undirected graph. Formally, let $V = \{\mathcal{S}_1, \ldots, \mathcal{S}_n\}$ denote the set of all surfaces in the model, and the edge set is defined as:
\begin{equation}
    E = \{(\mathcal{S}_i, \mathcal{S}_j) \mid \mathcal{S}_i \cap \mathcal{S}_j \neq \emptyset\}
\end{equation}
where $\mathcal{S}_i \cap \mathcal{S}_j$ represents the common boundary between two surfaces. This graph can be represented by an adjacency matrix $A \in \{0,1\}^{n \times n}$:
\begin{equation}
    A_{ij} = \begin{cases}
        1, & \text{if } (\mathcal{S}_i, \mathcal{S}_j) \in E \\
        0, & \text{otherwise}
    \end{cases}
\end{equation}

This graph structure preserves the face-adjacency topology of the B-Rep model, enabling graph neural networks to capture local neighborhood structures as well as global topological features through multi-layer message passing.

% %节点/面特征
\subsubsection{Geometric Features}

To effectively represent the geometric shapes of CAD surfaces for downstream deep learning tasks, we propose a compact and rigid-transformation-invariant feature extraction scheme. Many existing studies directly use the three-dimensional coordinates of sampled points on surfaces as features. Although intuitive, their main drawback is extreme  sensitivity to the spatial pose (rotation and translation) of the model, which leads to poor generalization of network models.

To address this issue, we employ curvature from differential geometry as the core geometric descriptor. Curvature is an intrinsic property of surfaces that describes only the bending behavior of the surface itself, remaining invariant to its position and orientation in space. This intrinsic invariance makes it well-suited for constructing robust geometric features.

Our feature extraction pipeline proceeds as follows. First, to ensure a fixed-dimensional representation for arbitrary surfaces, we perform uniform sampling in the normalized two-dimensional parametric (UV) domain (i.e., $\Omega = [0, 1] \times [0, 1]$), constructing a regular $n \times n$ grid, thereby obtaining $n^2$ regularly arranged sample points on the surface.

Second, we compute local curvature information at each of these $n^2$ sample points. Specifically, we calculate Gaussian curvature ($K$) and mean curvature ($H$). Gaussian curvature, defined as the product of the two principal curvatures, characterizes the intrinsic curvature of the surface and serves as a key indicator for distinguishing local shape types (e.g., planar, spherical, or saddle-shaped regions). Mean curvature, the arithmetic mean of the two principal curvatures, describes the average bending of the surface. These two curvature values can be computed from the first and second fundamental form coefficients ($E, F, G$ and $L, M, N$) of the surface:
\begin{equation}
    K = \frac{LN - M^2}{EG - F^2}, \quad H = \frac{EN - 2FM + GL}{2(EG - F^2)}
\end{equation}
For the $(i, j)$-th point in the grid, we compute its curvature pair $(K_{ij}, H_{ij})$.

Finally, we flatten and concatenate the curvature information of these $n^2$ points ($n \times n$) in a predetermined order (e.g., row-wise scanning) to form a final $2n^2$-dimensional feature vector $\mathbf{f}_{\mathcal{S}} \in \mathbb{R}^{2n^2}$:

\begin{equation}
    \mathbf{f}_{\mathcal{S}} = [K_{11}, K_{12}, \dots, K_{nn}, H_{11}, H_{12}, \dots, H_{nn}]^T
\end{equation}

This feature vector remains invariant to the spatial pose of the model and can more fundamentally characterize the geometric details of features such as fillets and chamfers, providing high-quality, robust, and discriminative input for subsequent recognition and analysis tasks.

\subsubsection{Attribute Features}

In addition to continuous geometric features (curvature) that rely on the intrinsic bending variations of surfaces, we introduce a set of discrete attributes to describe the macroscopic morphology of each surface in the CAD model. These attributes are carefully designed to capture geometric features with specific engineering semantics that are formed during manufacturing processes. We primarily extract two core attributes: face width and dihedral angle between adjacent faces.

\subsubsection{Face Width ($W_{\mathcal{S}}$)}

\begin{figure}[htbp]
    \centering
    \includegraphics[width=0.5\textwidth]{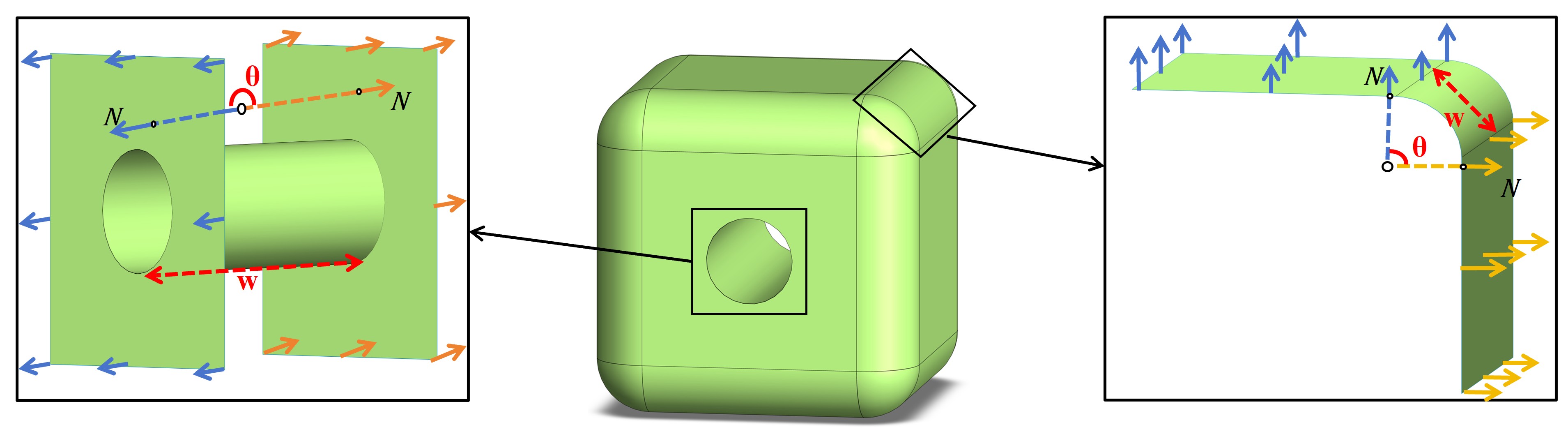} 
    \caption{Schematic diagram of face width $W_{\mathcal{S}}$ and dihedral angle $\theta_{\mathcal{S}}$ between adjacent faces.}
    \label{fig:attribute_info}
\end{figure}

Face width aims to quantify the characteristic dimension of a surface, which is crucial for distinguishing manufacturing features of different specifications. For a given surface $\mathcal{S}$, we first obtain the set of all boundary edges $\mathcal{E}$. Then, we identify the two longest boundary edges from it, denoted as $e_1$ and $e_2$. Let the midpoints of these two edges be $\bm{p}_1$ and $\bm{p}_2$, respectively. The face width $W_{\mathcal{S}}$ is defined as the Euclidean distance between these two midpoints:
\begin{equation}
    W_{\mathcal{S}} = \| \bm{p}_2 - \bm{p}_1 \|_2
\end{equation}
If the surface $\mathcal{S}$ has only one boundary edge, we set $W_{\mathcal{S}}$ to be the length of that edge. A schematic illustration of this definition is shown in Figure~\ref{fig:attribute_info}.

\subsubsection{Dihedral Angle $\theta_{\mathcal{S}}$}

The dihedral angle between adjacent faces describes the topological environment in which a surface resides, particularly the sharpness of the original edge or corner it replaces. The calculation of this attribute is also based on the two longest boundary edges $e_1$ and $e_2$. We search for two other surfaces adjacent to $e_1$ and $e_2$ respectively (excluding $\mathcal{S}$ itself), denoted as $\mathcal{S}'_1$ and $\mathcal{S}'_2$. The dihedral angle $\theta_{\mathcal{S}}$ is defined as the dihedral angle between these two adjacent surfaces $\mathcal{S}'_1$ and $\mathcal{S}'_2$. A schematic illustration of this definition is shown in Figure~\ref{fig:attribute_info}.

We define an important special case for this attribute: for self-closed surfaces such as cylinders, some of their boundary edges have the surface itself as their adjacent face. Under this topological structure, when $\mathcal{S}'_k = \mathcal{S}$, we define $\theta_{\mathcal{S}} = 0$. This definition aligns with engineering intuition, as it indicates that the boundary is not produced by a sharp intersection of two distinct faces, but rather represents a smooth self-continuation. Since the curvature information of cylindrical surfaces and filleted surfaces is similar, utilizing dihedral angle attributes can more effectively distinguish between these two types of geometric features.

\subsection{Network Architecture}
\label{sec:network_architecture}

The graph neural network architecture proposed in this paper is designed to efficiently fuse multi-source geometric information for feature classification. As illustrated in Figure~\ref{fig:overview}, the architecture integrates the model's topological structure with various geometric features, processing them through several key modules: Preprocessing Layers, Embedding Layers, a Graph Convolutional (GCN) Layer, step-wise Fusion Layers, and a final Classification Layer.

\subsubsection{Input Representation and Preprocessing}
The network takes two main types of input: the face adjacency graph, representing the topological structure, and three independent sets of node features: (1) local curvature features (Gaussian and Mean Curvature), (2) the face width attribute, and (3) the face angle attribute. The face adjacency graph is used directly to define the computational structure of the GCN. The three sets of node features are each passed through their own independent Preprocessing Layer (a fully connected layer with ReLU activation) to map them into a unified 32-dimensional feature space. Notably, the processed curvature features are further fed into an Embedding Layer that upsamples their feature dimension from 32 to 64, enhancing their representational capacity before the graph convolution step.

\subsubsection{Graph Convolutional Layers}
Subsequently, the dimensionally-enhanced curvature features serve as the initial node representations and are fed, along with the face adjacency graph, into a multi-layer Graph Convolutional Network (GCN) module \cite{kipf2016semi}. The GCN module consists of $N$ sequential GCN layers (default $N=3$), with each layer having an output dimension of 64. Each GCN layer learns high-level features that represent geometric and topological relationships by aggregating neighborhood information. The propagation rule for each layer is defined as:
\begin{align}
   H^{(l+1)} = \sigma\left(\tilde{D}^{-\frac{1}{2}}\tilde{A}\tilde{D}^{-\frac{1}{2}}H^{(l)}W^{(l)}\right)
\end{align}
where $\tilde{A} = A + I$ is the adjacency matrix with added self-loops, $\tilde{D}$ is the corresponding degree matrix, $W^{(l)}$ is a trainable weight matrix for the $l$-th layer, and $\sigma$ denotes the Tanh activation function.

\subsubsection{Step-wise Feature Fusion}
A key feature of this architecture is its step-wise feature fusion design, which progressively injects geometric information from different sources into the network. To achieve this, the output of the GCN layer ($N\times64$) is first passed through an Embedding Layer that downsamples its features to 32 dimensions. The network then performs a two-step fusion process:
\begin{enumerate}
    \item The downsampled features are concatenated with the preprocessed face width features and fed into the first Fusion Layer.
    \item The output from the previous step is then concatenated with the preprocessed face angle features and fed into the second Fusion Layer.
\end{enumerate}
Each fusion layer consists of a linear transformation, Batch Normalization, ReLU activation, and Dropout. This hierarchical fusion strategy allows the model to deeply integrate the global topological context learned by the GCN with specific local geometric attributes.

\subsubsection{Classification Layer}
Finally, the Classification Layer, composed of a fully connected layer and a Softmax activation function, maps the final fused feature vectors to a probability distribution over the output classes (e.g., fillet or non-fillet), yielding the final classification result for each face.

\subsubsection{Activation Functions and Regularization}
To ensure stable training and model robustness, specific activation functions and regularization techniques are employed. The GCN layer uses a Tanh activation function, while the final classification layer uses Softmax. All other layers, including the preprocessing, embedding, and fusion layers, utilize the ReLU activation function. Batch Normalization is applied to the intermediate layers of the network (such as the embedding, GCN, and fusion layers) to stabilize the learning process. Furthermore, Dropout is used in the GCN layer, the downsampling embedding layer, and both fusion layers to mitigate overfitting and enhance model generalization.

\section{Dataset}

To address the limited feature types and simple geometric structures in existing machining feature datasets, Colligan et al.\cite{colligan2022hierarchical} proposed the MFCAD++ dataset creation tool. This method first generates a rectangular blank with xyz dimensions randomly selected within the range of 10-50mm. Subsequently, 24 types of machining features and blank faces are divided into five groups: steps, slots, through features, non-through features, and transition features. Features are then generated in random combination sequences following the manufacturing logic of outside-to-inside and through-to-non-through. For each feature in the sequence, the method first triangulates the CAD model surface, randomly selects sampling points within the facets to generate bounding boxes conforming to four sketch types, then filters valid bounding boxes through ray casting and solid integrity verification while filtering out concave edges to avoid feature semantic loss. Finally, machining features are added at the bounding box locations and class labels are assigned to modified or newly created faces.

\begin{figure}[htbp]
    \centering
    \subfloat[]{
        \includegraphics[width=0.15\textwidth]{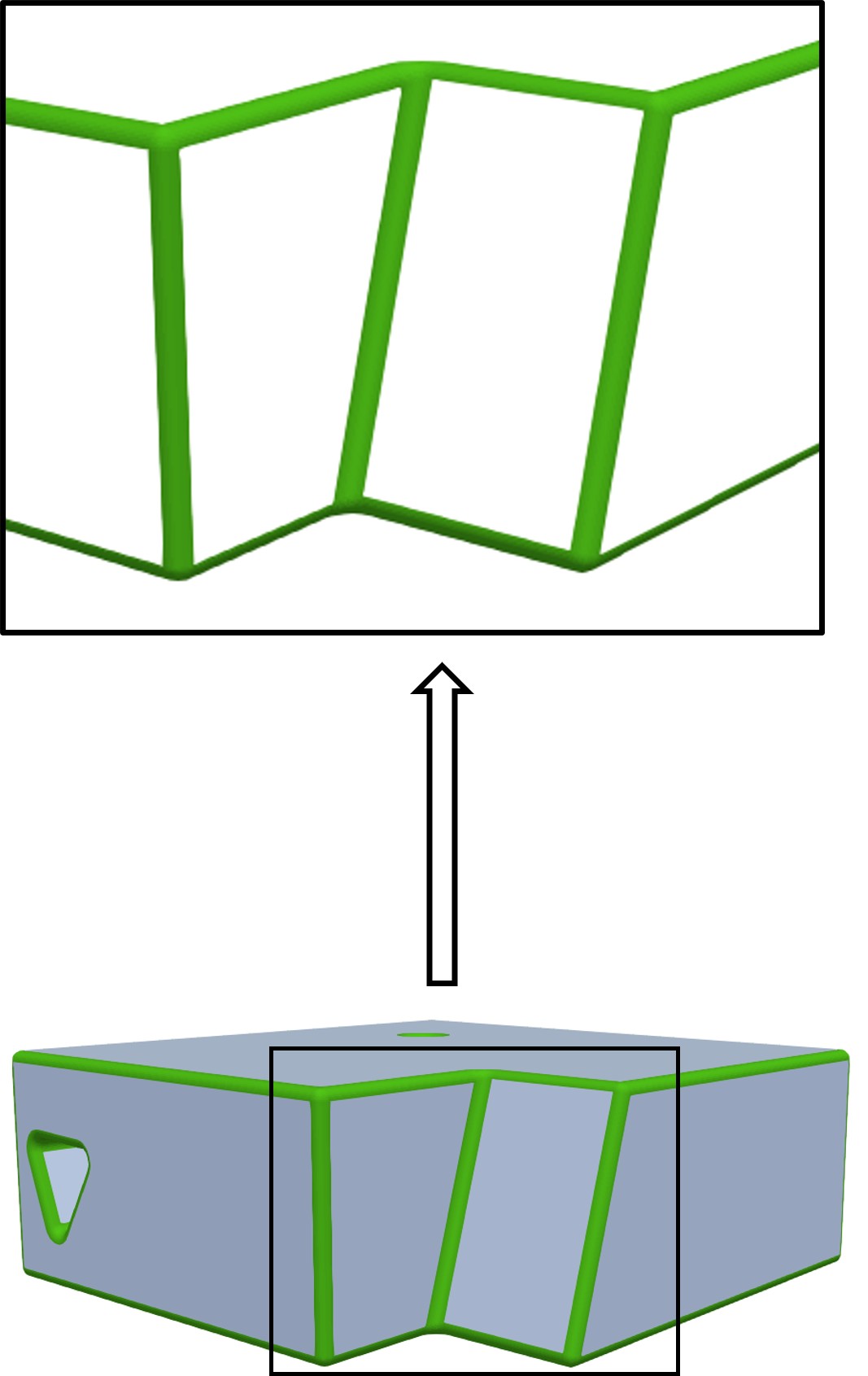}
        \label{fig:uniform_fillet}
    }
    \subfloat[]{
        \includegraphics[width=0.155\textwidth]{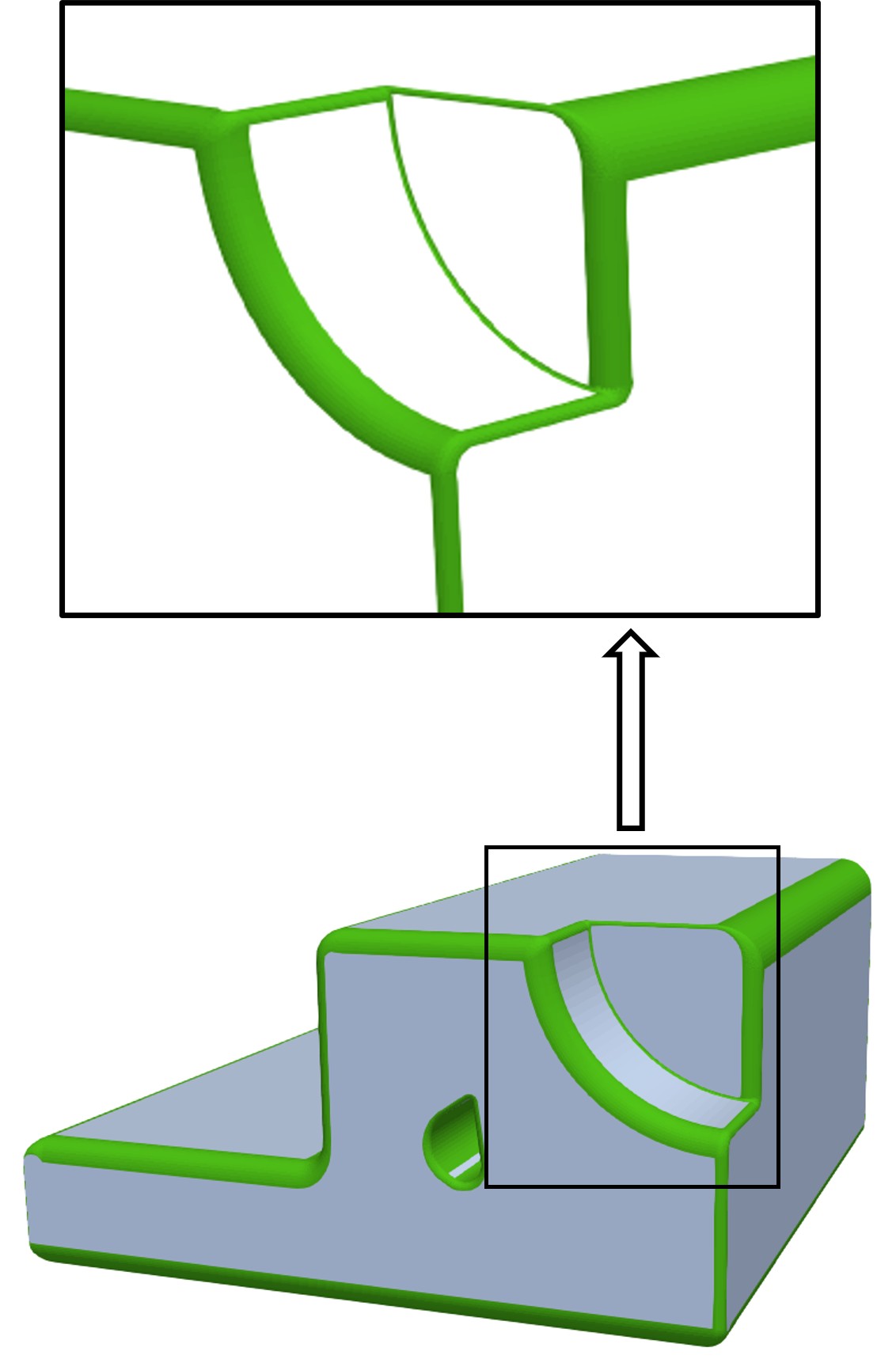}
        \label{fig:variable_fillet}
    }
    \subfloat[]{
        \includegraphics[width=0.15\textwidth]{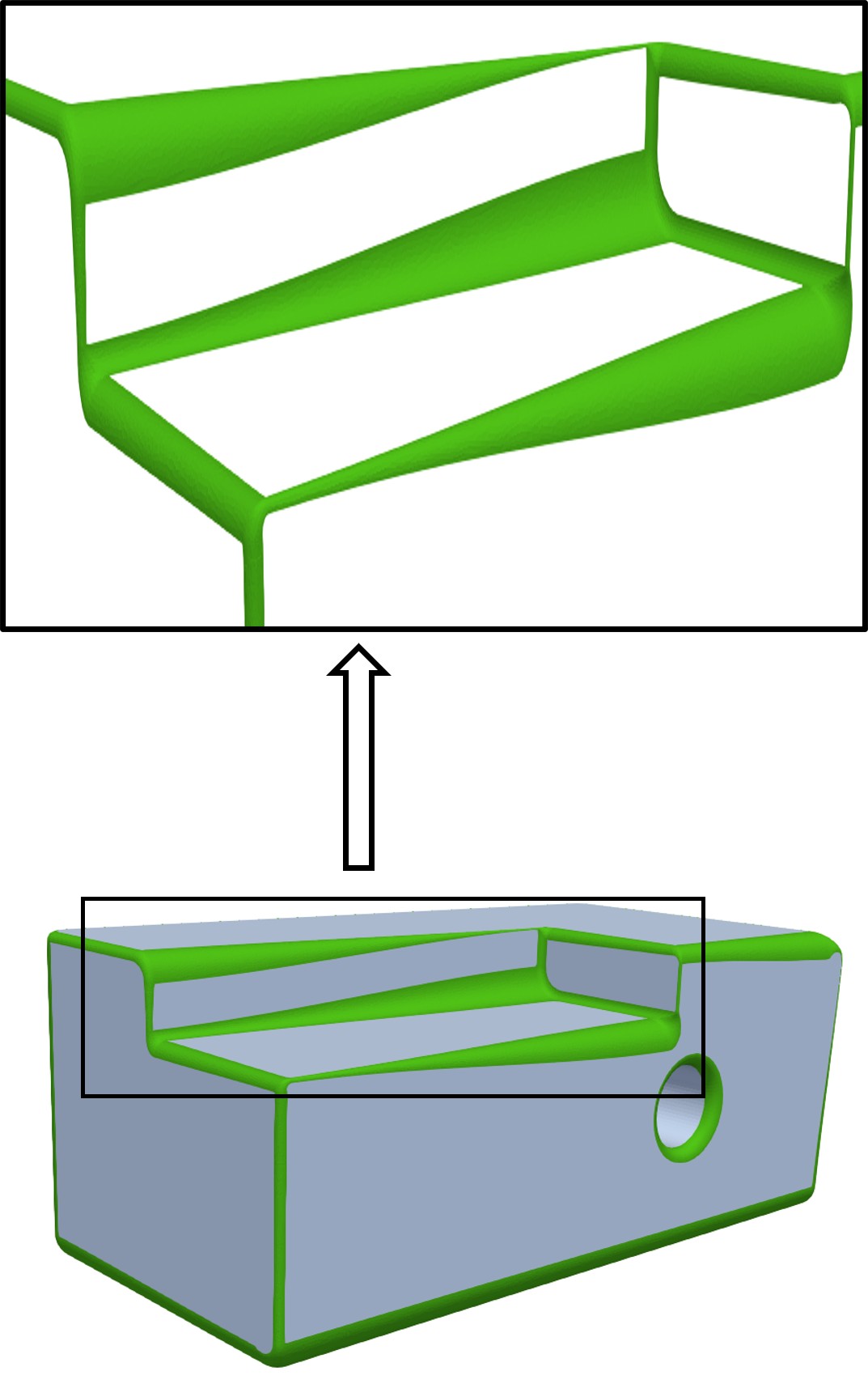}
        \label{fig:irregular_fillet}
    }
    \caption{Classification of fillet features in the dataset. (a) Uniform-radius regular fillets. (b) Variable-radius regular fillets. (c) Irregular fillets.}
    \label{fig:fillet_types}
\end{figure}

Although most existing datasets already contain fillet features, their fillet features are limited in quantity and type, particularly lacking complex geometric structures such as variable-radius fillets and continuous fillets, making it difficult to adequately cover the diverse design requirements in practical engineering. To address this, we construct a fillet-centric dataset based on the MFCAD++ dataset creation framework. This dataset not only contains a larger number and greater variety of fillet samples, but also introduces numerous fillet features with complex structures to better support application scenarios requiring high-precision modeling and strong generalization capability. According to the creation method, we categorize fillets into three types, as shown in Figure~\ref{fig:fillet_types}:

\begin{itemize}
   \item \textbf{Uniform-radius regular fillets}: All fillet features on a single CAD model have the same fixed radius;
   \item \textbf{Variable-radius regular fillets}: Fillet features on a single CAD model can have different fixed radius;
   \item \textbf{Irregular fillets}: A single CAD model contains both variable-radius fillets and fixed-radius fillets, with fillet radius
   randomly generated.
\end{itemize}

\begin{figure*}[htbp]
    \centering
    \includegraphics[width=0.95\textwidth]{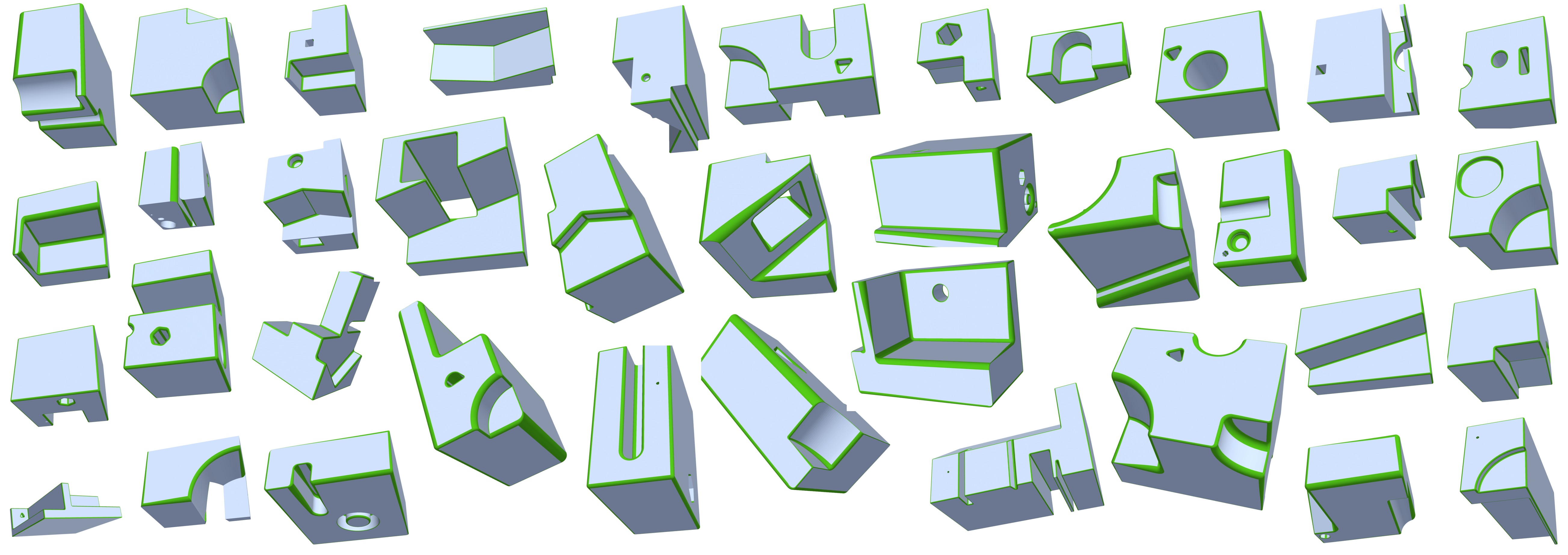} 
    \caption{Representative samples from the 4,486 models.}
    \label{fig:dataset}
\end{figure*}

The dataset comprises 4,486 STEP files, with representative samples shown in Figure~\ref{fig:dataset}. Face labels are categorized into two classes: non-fillet faces and fillet faces. The dataset is split into training, validation, and test sets with a ratio of 8:1:1 by the number of models.

\section{Experiments}

\subsection{Comparison with Baselines}

\subsubsection{Evaluation Metrics}
\label{sec:metrics}

To quantitatively evaluate the classification performance of our proposed model, we employ four standard metrics derived from the confusion matrix: Accuracy, Precision, Recall, and F1-score. These metrics are calculated from the counts of True Positives (TP), True Negatives (TN), False Positives (FP), and False Negatives (FN), with their definitions as follows:

\begin{align}
    \text{Accuracy} &= \frac{\text{TP} + \text{TN}}{\text{TP} + \text{TN} + \text{FP} + \text{FN}} \\
    \text{Precision} &= \frac{\text{TP}}{\text{TP} + \text{FP}} \\
    \text{Recall} &= \frac{\text{TP}}{\text{TP} + \text{FN}} \\
    \text{F1-score} &= \frac{2 \cdot \text{Precision} \cdot \text{Recall}}{\text{Precision} + \text{Recall}}
\end{align}

In our binary classification task, we define fillet faces as the positive class; therefore, TP, FP, and FN are computed based on the recognition results for fillet faces. The specific role of each metric is detailed below:
\begin{itemize}
    \item \textbf{Accuracy:} Measures the overall proportion of correctly classified faces. While intuitive, this metric can be misleading on datasets with an imbalanced class distribution.
    \item \textbf{Precision:} Measures the proportion of faces predicted as fillets that are actually fillets. It focuses on the exactness of the positive predictions, thereby penalizing False Positives.
    \item \textbf{Recall:} Measures the proportion of all actual fillet faces that were successfully identified by the model. It assesses the model's ability to find all positive instances, thereby penalizing False Negatives.
    \item \textbf{F1-score:} The harmonic mean of Precision and Recall. It provides a single, balanced metric that is particularly important in cases of uneven class distribution, as it accounts for both False Positives and False Negatives.
\end{itemize}

The combined use of these metrics provides a comprehensive and objective assessment of the model's performance from multiple perspectives.

\subsubsection{Experimental Setup}
\label{sec:experimental_setup}

To ensure the reproducibility and fairness of our experiments, this section details the specific implementation of the FilletRec model, its training procedure, and the configuration for comparison against baseline models.

\textbf{FilletRec Model Configuration:} 
The FilletRec model was implemented using the TensorFlow framework, with all training and testing conducted on a workstation equipped with a 12th Gen Intel\textsuperscript{\textregistered} Core\textsuperscript{\texttrademark} i7-12700KF CPU @ 3.60 GHz. For the input features, the Gaussian and mean curvatures of each surface were sampled on a $5 \times 5$ parametric grid. Regarding the network architecture, the model is designed for a binary classification task. The preprocessing layer has an output dimension of 32. The graph convolutional module consists of 3 layers. Weights were initialized using a truncated normal distribution. For optimization, we employed the Adam optimizer \cite{zhang2018improved} with an initial learning rate of 1e-3 and a weight decay of 1e-4. To enhance the model's generalization capability, we incorporated L2 regularization (with a coefficient of 0.0005) and a Dropout strategy with a rate of 0.3.

\textbf{Baseline Models and Training Protocol:} 
To comprehensively evaluate the performance of FilletRec, we selected two state-of-the-art CAD feature recognition models, Hierarchical CADNet \cite{colligan2022hierarchical} and AAGNet \cite{wu2024aagnet}, as baselines. The specific configurations were as follows:
\begin{itemize}
    \item \textbf{Hierarchical CADNet\footnote{\url{https://github.com/xupeiwust/hierarchical-cadnet}}:} Based on the ablation study in its original paper \cite{colligan2022hierarchical}, we selected the Hierarchical CADNet(Edge) variant, which incorporates edge convexity information, as it demonstrated superior feature recognition capabilities compared to the version relying solely on face adjacency.
    \item \textbf{AAGNet\footnote{\url{https://github.com/whjdark/AAGNet.git}}:} Since our dataset only contains face-level semantic labels, we activated and trained only the semantic segmentation head of the AAGNet model\cite{wu2024aagnet}.
\end{itemize}

For a fair comparison, all models (FilletRec, Hierarchical CADNet, and AAGNet) were trained for 30 epochs on our novel fillet dataset. We adapted the final classification layer of the baseline models to have 2 output classes, aligning with our binary classification task. All other parameters for the baseline models were kept at their default settings as specified in their original publications.

\subsubsection{Experimental Results}

This section presents the quantitative comparison results of FilletRec against two advanced baseline models on our dataset, with detailed data provided in Table~\ref{tab:performance_comparison}.

\begin{table}[h]
\centering
\small
\caption{Comparative results on our dataset. Precision, F1 and Recall metrics are only for fillet feature.}
\resizebox{\columnwidth}{!}{%
\begin{tabular}{cccccc}
\toprule
\textbf{Method} & \textbf{avg Acc (\%)} & \textbf{Precision (\%)} & \textbf{F1 (\%)} & \textbf{Recall (\%)} & \textbf{Parameters (M)} \\
\midrule
Hierarchical CADNet & 99.47 & 99.92 & 99.54 & 99.16 & 9.74 \\
AAGNet & 99.96 & 99.98 & 99.97 & 99.97 & 0.37\\
FilletRec & 99.91 & 99.91 & 99.93 & 99.94 & 0.02 \\
\bottomrule
\end{tabular}
}
\vspace{2mm}
\label{tab:performance_comparison}
\end{table}

Our proposed FilletRec model achieves highly competitive performance, reaching an overall accuracy of \textbf{99.91\%}. Concurrently, its precision, F1-score, and recall for the fillet feature class all exceed 99.9\%.  Compared to the baseline models, FilletRec achieves performance nearly equal to AAGNet (99.96\%) and significantly superior to Hierarchical CADNet (99.47\%), while maintaining significantly reduced model complexity.

The primary advantage of FilletRec lies in its exceptional model efficiency. The model comprises only \textbf{0.02M} parameters, a scale significantly smaller than Hierarchical CADNet (9.74M) and AAGNet (0.37M). Specifically, FilletRec's parameter count is merely 5.4\% of AAGNet's and 0.2\% of Hierarchical CADNet's. This result demonstrates that FilletRec strikes an excellent balance between recognition accuracy and model complexity, delivering performance comparable to state-of-the-art methods at a remarkably low computational cost. This highlights its efficient architectural design and its significant potential for practical applications.

\begin{figure*}[htbp]
    \centering
    \includegraphics[width=0.85\textwidth]{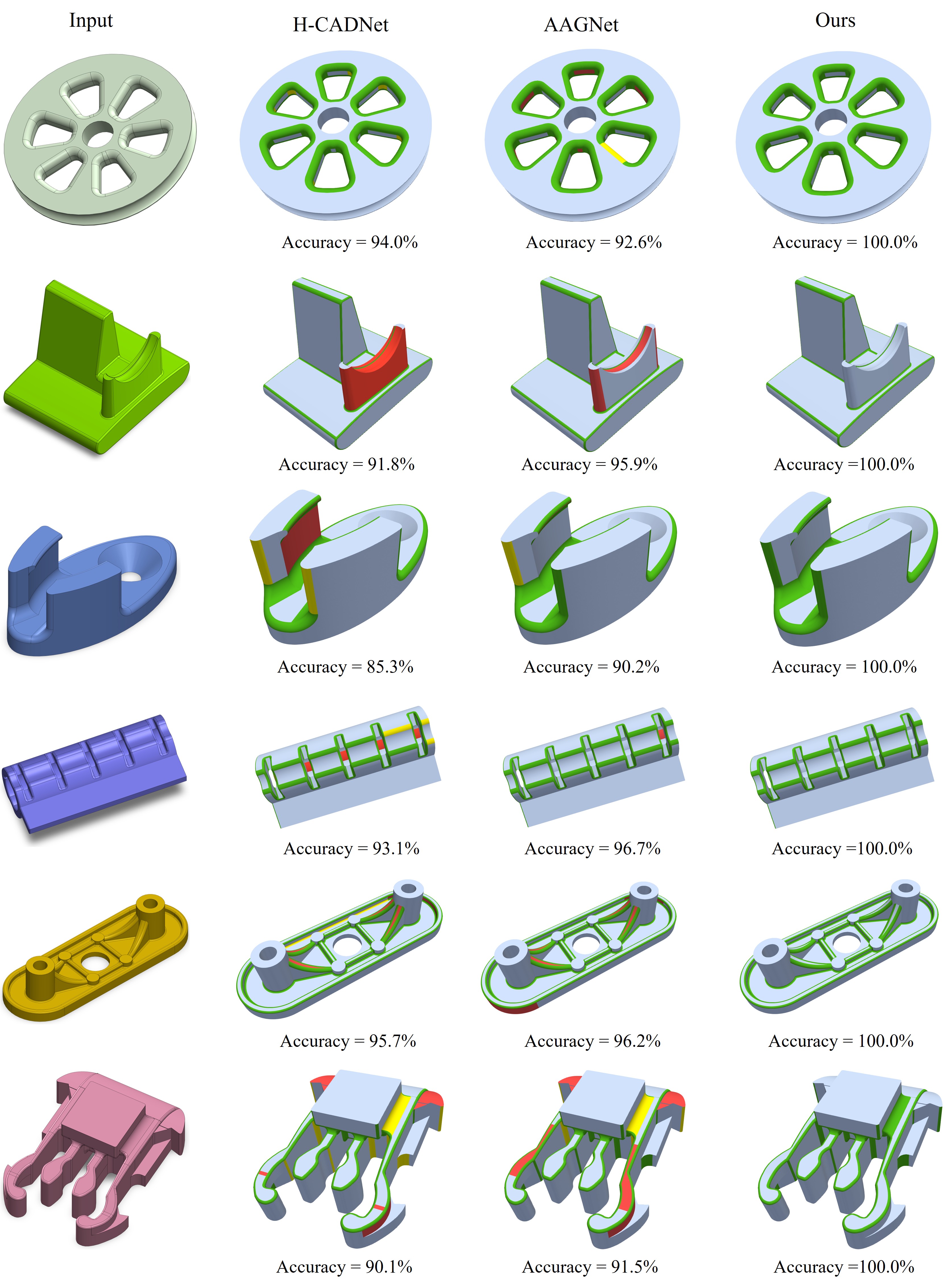} 
    \caption{Qualitative comparison of fillet recognition on complex models from the ABC dataset. Green indicates correctly identified fillets (TP), yellow indicates missed fillets (FN), and red denotes misidentified non-fillet faces (FP). Our method (FilletRec) consistently outperforms the baseline models, demonstrating superior generalization capabilities.}
    \label{fig:compare_AI_OURS}
\end{figure*}

In addition to the quantitative evaluation on our custom dataset, we conducted a more challenging generalization test to highlight the advantages of FilletRec. For this, we randomly selected six representative and complex models from the ABC dataset \cite{koch2019abc}. Figure~\ref{fig:compare_AI_OURS} shows the comparison results, where green represents correctly identified fillets (TP), yellow indicates missed fillets (FN), and red denotes misidentified non-fillet faces (FP).

The limitations of the baseline models are clearly revealed in these cases. The performance of Hierarchical CADNet was the most unstable, with its accuracy dropping from 95.7\% to as low as 85.3\% (third row). Although AAGNet's performance was better, its accuracy still fluctuated between 90.2\% and 96.7\%. In stark contrast, FilletRec achieved a perfect 100\% accuracy on all six complex test models.

Overall, on this challenging set of samples, FilletRec's average accuracy reached 100\%, significantly higher than AAGNet's (93.9\%) and Hierarchical CADNet's (91.7\%). This not only validates our model's strong generalization capability but also demonstrates its great potential as a reliable CAD feature recognition tool in complex scenarios.

\subsection{Dataset Comparison}

To validate the challenging nature of our proposed dataset for evaluating model generalization, we conducted a cross-dataset comparative experiment. First, we trained two baseline models, Hierarchical CADNet and AAGNet, on the standard MFCAD++ dataset. They achieved average accuracies of 96.55\% and 99.23\% respectively, which is largely consistent with the results reported in their original papers, thereby validating the fidelity of our experimental setup. We then evaluated these validated, pre-trained models on both the MFCAD++ test set and our new dataset. The detailed performance metrics for fillet recognition are summarized in Table~\ref{tab:cross_dataset_comparison}.

As shown in the Table~\ref{tab:cross_dataset_comparison}, both models perform well when tested on the MFCAD++ dataset, achieving high F1-scores of 97.00\% and 99.66\%, respectively. However, a stark contrast is observed when these same models are evaluated on our dataset. Their performance degrades significantly, especially for AAGNet, despite its superior performance on the original benchmark. Its recall for fillet features drops sharply from 99.75\% (on MFCAD++) to just 48.58\% (on our dataset), causing its F1-score to fall sharply from 99.66\% to 65.38\%.

This significant performance gap clearly shows that the existing models have likely overfitted to the distribution of the training data. By including more diverse and complex fillet instances that represent industrial scenarios, our dataset more effectively exposes the generalization weaknesses of existing models when confronted with out-of-distribution samples. Therefore, our dataset provides a more reliable and challenging benchmark for evaluating and advancing research on the robustness and generalization of CAD feature recognition models.

\begin{table*}[h]
\centering
\small
\caption{Performance comparison of models trained on MFCAD++ and tested on different datasets. This cross-dataset evaluation highlights the generalization challenge posed by our dataset. All metrics shown are for fillet features recognition.}
\label{tab:cross_dataset_comparison}
\begin{tabular}{l c ccc ccc}
\toprule
\multirow{2}{*}{\textbf{Network}} & \multirow{2}{*}{\textbf{Params. (M)}} & \multicolumn{3}{c}{\textbf{Test on MFCAD++}} & \multicolumn{3}{c}{\textbf{Test on Our Dataset}} \\
\cmidrule(lr){3-5} \cmidrule(lr){6-8}
& & Precision (\%) & F1 (\%) & Recall (\%) & Precision (\%) & F1 (\%) & Recall (\%) \\
\midrule
Hierarchical CADNet & 9.76 & 98.30 & 97.00 & 95.73 & 99.37 & 91.53 & 84.83 \\
AAGNet              & 0.38 & 99.58 & 99.66 & 99.75 & 99.94 & 65.38 & 48.58 \\
\bottomrule
\end{tabular}
\end{table*}

To provide further qualitative evidence of our dataset's ability to enhance model generalization, we designed a critical comparative test. We deployed Hierarchical CADNet and AAGNet models, which were trained separately on the MFCAD++ dataset and our own, onto an identical set of test cases featuring complex topological structures.

\begin{figure*}[htbp]
    \centering
    \includegraphics[width=0.9\textwidth]{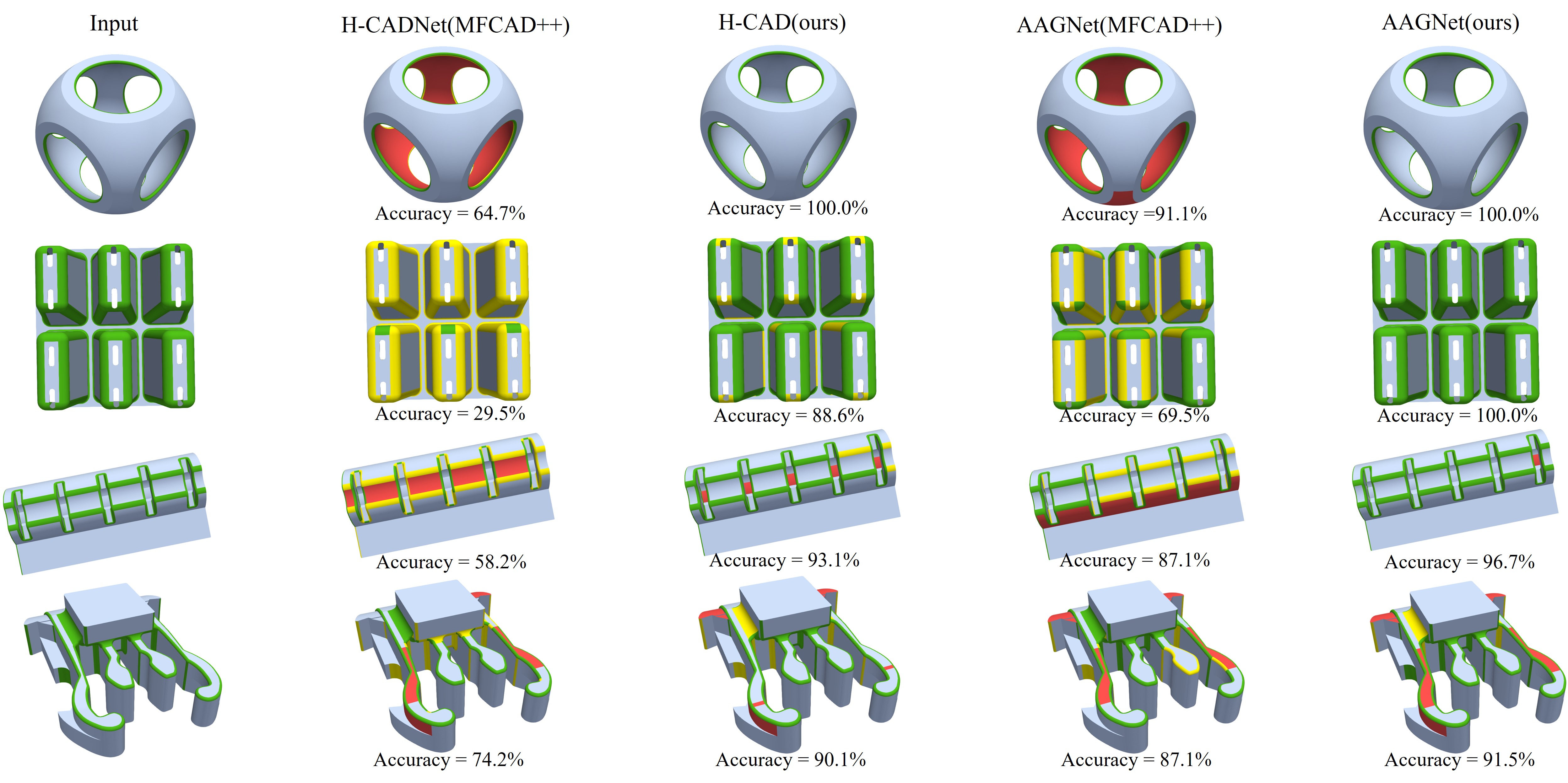} 
    \caption{Impact of training data on model generalization. Each row shows a complex test case. The columns compare the performance of baseline models when trained on the standard MFCAD++ dataset versus our proposed dataset. Models trained on our dataset exhibit significantly improved robustness and accuracy.}
    \label{fig:diff_dataset_res}
\end{figure*}

The results, illustrated in Figure~\ref{fig:diff_dataset_res} (where each row represents a distinct test case), reveal the decisive impact of training data quality on final model performance. In every instance, the models trained using our dataset exhibited significantly better performance. This improvement is especially significant for the test case in the second row, where the accuracy of Hierarchical CADNet surged from 29.5\% (when trained on MFCAD++) to 88.6\% (when trained on our dataset). Similarly, for the first-row case, AAGNet's performance improved from 91.1\% to a perfect 100\% recognition.

This consistent and substantial performance enhancement provides clear evidence that our dataset, owing to its greater diversity and complexity, guides models to learn more robust and fundamental geometric and topological representations of fillet features. This, in turn, equips them with the ability to handle complex scenario capability that is difficult to fully achieve when training solely on existing benchmarks such as MFCAD++.

\subsection{Ablation Experiment}

To quantitatively assess the contribution of each input feature to the recognition performance of the FilletRec model, we designed a series of ablation studies. By systematically removing key information—namely Gaussian curvature, mean curvature, face width, and dihedral angle—we could precisely measure the impact of each component. The results are detailed in Table~\ref{tab:ablation_study}.

\begin{table}[htbp]
\centering
\caption{Ablation study on the impact of different input features on the performance of the FilletRec model.}
\begin{tabular}{cc}
\toprule
\textbf{Network version} & \textbf{avg Accuracy (\%)} \\
\midrule
FilletRec & 99.91 \\
no gaussian curvature attr. & 99.74 \\
no mean curvature attr. & 94.87 \\
no curvature attr. & 93.46 \\
no surface width attr. & 99.51 \\
no adjacent faces' angle attr. & 99.42 \\
\bottomrule
\end{tabular}
\label{tab:ablation_study}
\end{table}

\begin{figure}[htbp]
    \centering
    \includegraphics[width=0.48\textwidth]{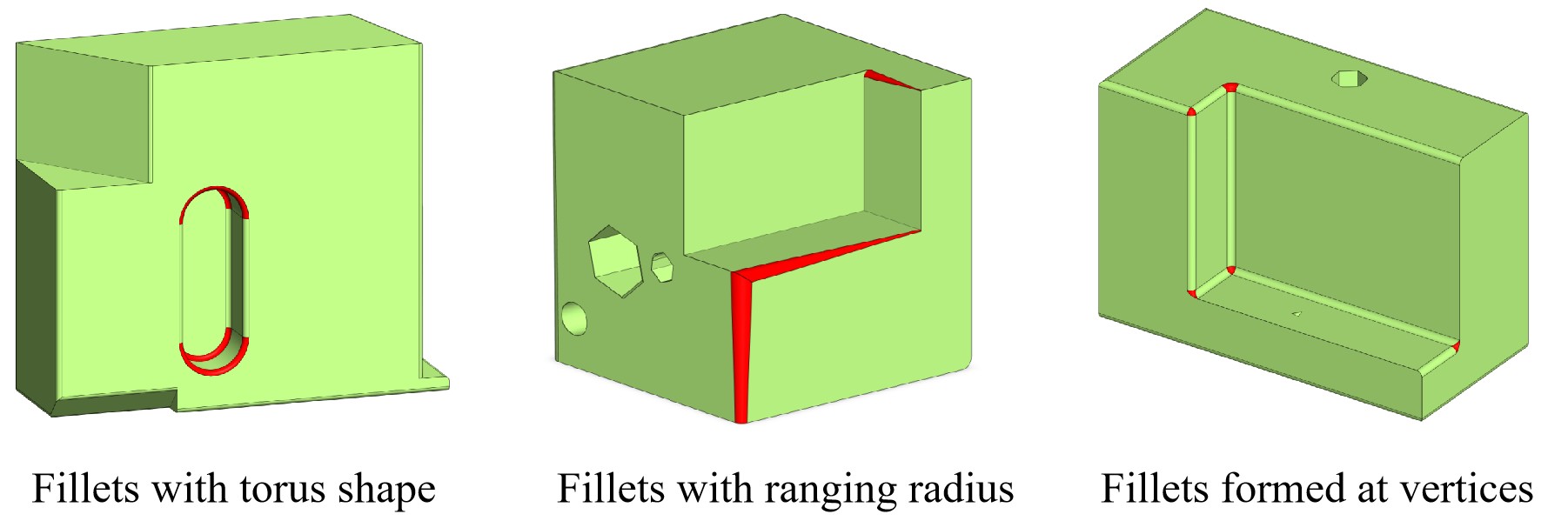} 
    \caption{Examples of complex fillet topologies with non-zero Gaussian curvature. These special cases are highlighted in red.}
    \label{fig:non_zero_gaussian_curvature}
\end{figure}

The experimental results show that the full model (Baseline), which integrates all features, achieves a benchmark accuracy of 99.91\%. When mean curvature was removed individually, the model's performance experienced a drastic decline, with accuracy dropping sharply to 94.87\%. This highlights its decisive role as a core feature. In contrast, removing Gaussian curvature only led to a slight accuracy drop to 99.74\%. This difference occurs because most surfaces in our dataset are developable surfaces, for which the Gaussian curvature is theoretically zero. Only a few complex fillet topologies, such as those with toroidal shapes, variable radius, or at vertex fillets, exhibit non-zero Gaussian curvature, as illustrated in Figure \ref{fig:non_zero_gaussian_curvature}. A feature composed predominantly of zero values struggles to provide effective gradients during neural network training, thus limiting its discriminative power. Conversely, mean curvature consistently reflects the local degree of surface bending, providing the model with more discriminative geometric information. When both types of curvature information were removed simultaneously, the accuracy fell to its lowest point of 93.46\%, further confirming that curvature is indispensable prior knowledge for the fillet recognition task.

On the other hand, removing the geometric attributes of face width and dihedral angle resulted in marginal accuracy drops to 99.51\% and 99.42\%, respectively. This indicates that while they are not the determining factors of performance, they still serve as effective auxiliary roles, providing beneficial supplementary information for the model's decision-making process.

In summary, this ablation study clearly reveals the key to the FilletRec model's success: it is driven by mean curvature as its core, and enhanced by combining multi-dimensional geometric information including Gaussian curvature, face width, and dihedral angle. This carefully designed combination of complementary features ensures 
the model's exceptional recognition performance and robustness.

\subsection{Comparison with Traditional Methods}

\begin{figure*}[htbp]
    \centering
    \includegraphics[width=0.8\textwidth]{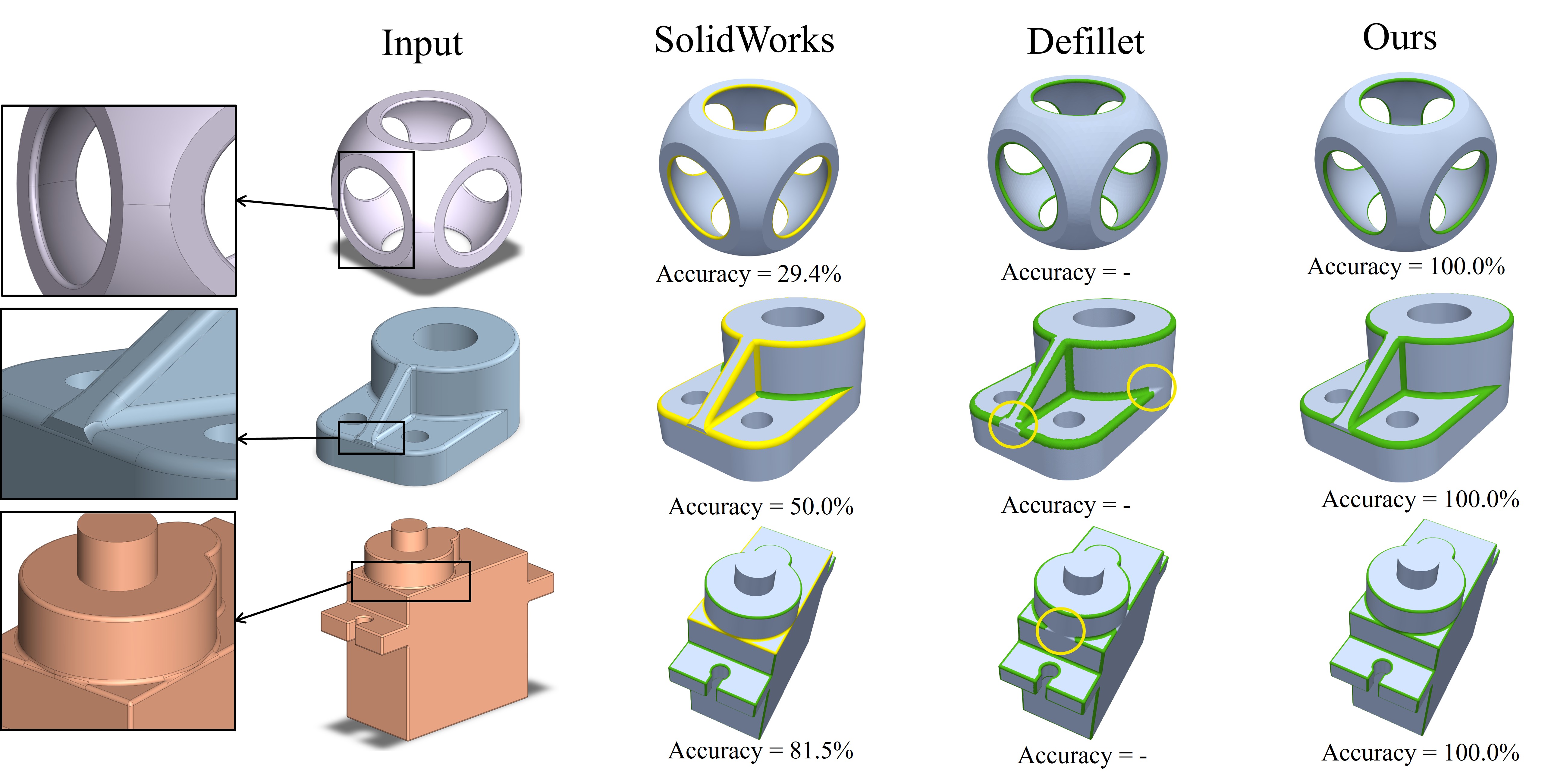} 
    \caption{Comparison of fillet recognition results against commercial software (SolidWorks) and a non-learning-based method (Defillet). While existing methods either under-segment (SolidWorks, yellow) or miss features (Defillet, circled errors), our data-driven approach (Ours) achieves complete and accurate recognition on all three complex models.}
    \label{fig:SolidWorks}
\end{figure*}

To evaluate the performance of FilletRec in the task of fillet recognition, we conducted a comparative study against the commercial CAD software SolidWorks \cite{matsson2025introduction} and a recent non-learning-based method, Defillet \cite{jiangDeFilletDetectionRemoval2025}, with the results illustrated in Figure \ref{fig:SolidWorks}.

The results demonstrate that non-learning-based methods still exhibit significant limitations in identifying complex fillet features. SolidWorks' built-in automatic recognition functionality performs poorly when faced with complex topological structures, especially when handling fillets at multi-surface junctions or those with variable radii.

While Defillet shows significant improvements over SolidWorks, it still suffers from missed detections for irregular fillets. Moreover, the method imposes stringent requirements on the quality of the input mesh; its heavy reliance on mesh integrity and uniformity severely restricts its applicability in practical engineering scenarios.

In contrast, our data-driven model FilletRec significantly outperforms these traditional methods. By learning from data, FilletRec can automatically recognize various types of fillets, including complex and irregular ones. The model maintains high accuracy and reliability even on challenging cases, demonstrating clear advantages over rule-based approaches.

\section{Fillet Removal}
\label{sec:fillet_removal}

\subsection{Overview}
\label{sssec:overview}

The removal of small or redundant features, a process known as defeaturing, is a long-standing challenge in preparing CAD models for CAE analysis \cite{thakur2009survey}. Our fillet removal method aims to robustly convert CAD models with smooth fillet transitions into geometric representations with sharp edges. Directly deleting the fillet faces creates complex holes on the model, and directly filling these holes is prone to generating low-quality meshes or even topological errors.

\begin{figure*}[htb]
    \centering
    \includegraphics[width=0.9\textwidth]{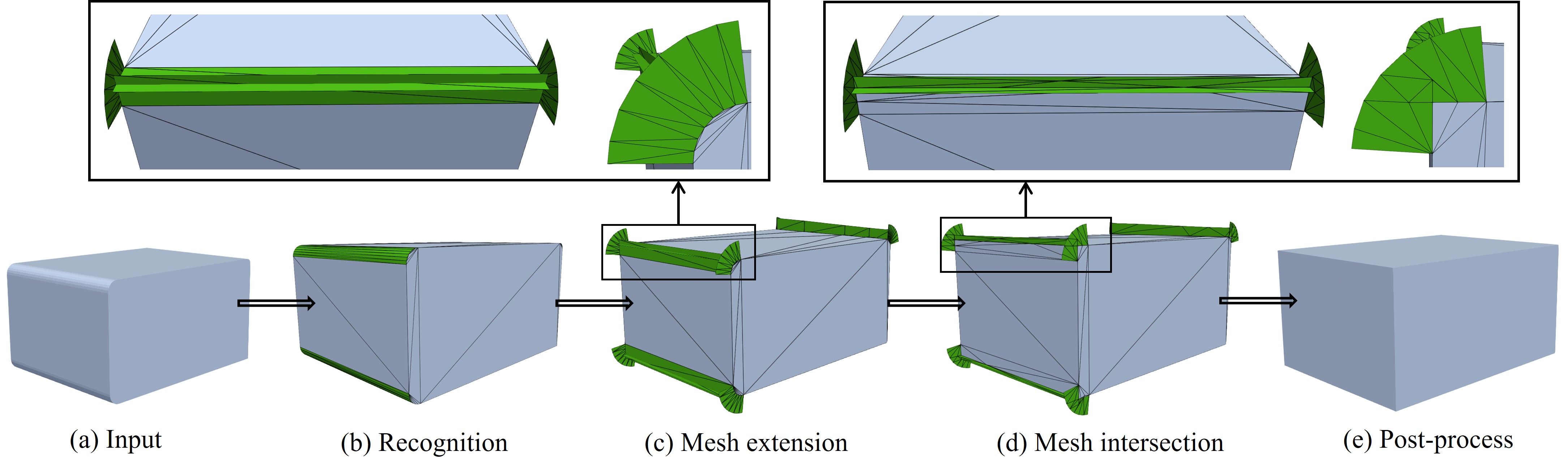} 
    \caption{Pipeline of our proposed fillet removal method.}
    \label{fig:defillet_overview}
\end{figure*}

To address this challenge, we propose a three-stage core strategy: "Extend-Intersect-Clean," whose pipeline is illustrated in Figure~\ref{fig:defillet_overview}. The core idea of this strategy is as follows: first, we use a deep learning network to accurately identify and remove all fillet faces. Then, instead of directly filling the resulting holes, we proactively extend the boundary mesh of the holes outwards, causing them to overlap, thereby constructing an intermediate state with self-intersections. Subsequently, a robust mesh intersection algorithm is used to accurately compute the sharp edges formed by the intersections of these extended patches. Finally, through a series of post-processing operations, we remove all redundant and floating mesh elements to produce a closed, manifold model. This approach effectively transforms the complex problem of hole filling into a more well-established problem of mesh Boolean operations, thereby significantly enhancing the robustness of the algorithm and the quality of the final model.

\subsection{Boundary Mesh Extension}
\label{sssec:extension}
After identifying and removing all fillet faces, holes that require repair are formed on the model. We process the remaining faces differently based on whether they are adjacent to the hole boundaries. For internal geometric surfaces not adjacent to any boundary, we perform a standard triangulation directly.

\begin{figure}[h]
\centering
\includegraphics[width=0.5\textwidth]{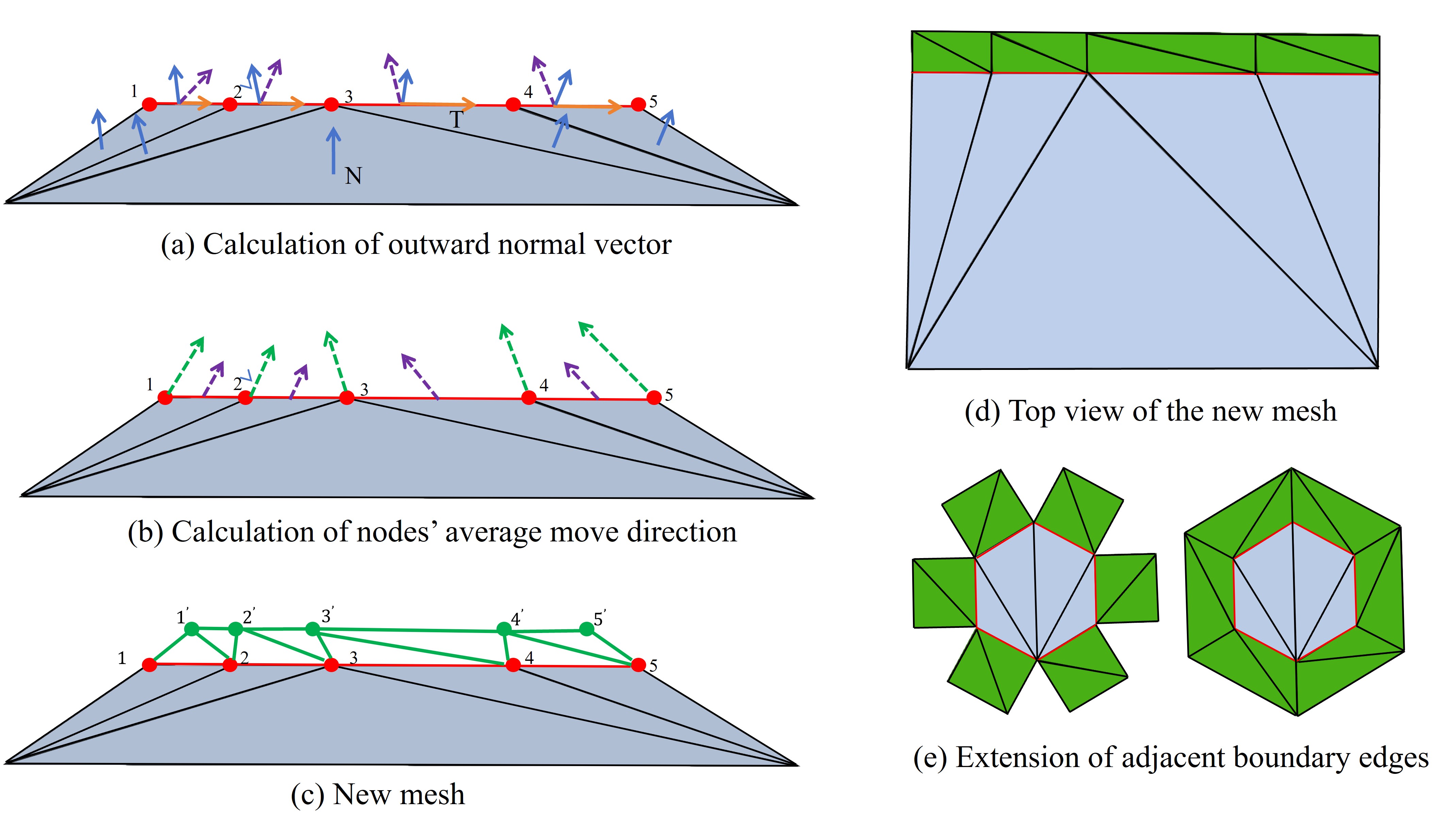}
\caption{Detailed illustration of the boundary mesh extension algorithm. (a) An outward normal vector (purple) is computed for each boundary segment using the adjacent face normal (N) and the edge tangent (T). (b) The final offset direction for each vertex (green, dashed) is calculated by averaging the normals of its neighboring segments. (c-d) New vertices are created along the offset directions, forming a new strip of triangular mesh (green) that extends the boundary. (e) The algorithm handles closed loops by processing adjacent edges concurrently to ensure a continuous, gap-free extension.}
\label{fig:extension_detail }
\end{figure}

For geometric surfaces adjacent to the hole boundaries, we employ a mesh extension algorithm to generate transitional geometry that covers the original fillet region. The process is illustrated in detail in Figure~\ref{fig:extension_detail }. The algorithm first triangulates these boundary-adjacent surfaces. Subsequently, for each vertex on the identified boundary loops, we compute a precise offset direction. As shown in Figure~\ref{fig:extension_detail }(a), this is achieved by first calculating an outward normal vector for each boundary segment, derived from the adjacent face normal (N) and the boundary edge tangent (T). To ensure a smooth and robust extension, the final offset direction for each vertex is determined by averaging the outward normals of its neighboring segments (Figure~\ref{fig:extension_detail }(b)).

A new vertex is then generated by offsetting the original vertex along this averaged direction. Using the original boundary vertices and their newly generated counterparts, we construct a strip of new triangular faces that forms the extended mesh (Figure~\ref{fig:extension_detail }(c-d)). To ensure the topological continuity of the extended mesh, especially at corners or on closed loops, adjacent boundary edges are processed concurrently to prevent the formation of gaps (Figure~\ref{fig:extension_detail }(e)).

Finally, by re-indexing the vertices, we merge the original mesh with the extended mesh to create a composite mesh containing self-intersections, which serves as the foundation for the subsequent intersection operation.

\subsection{Self-Intersection Resolution and Post-processing}
\label{sssec:postprocess}
The merged composite mesh contains the intended geometric intersections. To resolve these intersections and form sharp edges, we apply the robust mesh intersection algorithm proposed by Liu et al.~\cite{liuRobustFastLocal2026}, which can effectively handle complex self-intersection cases and generate correct intersection lines.

After the intersection operation, the resulting mesh may still contain invalid geometric elements. Therefore, we designed a post-processing pipeline to ensure the integrity and manifold property of the output model. This process includes two main steps:
\begin{enumerate}
    \item Open Boundary Cleaning: We iteratively remove all boundary triangles (i.e., triangles with at least one non-manifold edge) until no open boundaries remain on the model.
    \item Isolated Component Removal: Based on connectivity analysis, we identify all independent connected components in the mesh and retain only the largest main mesh body by volume, removing all other small, floating mesh fragments.
\end{enumerate}
Through this process, we finally obtain a structurally complete and clean watertight mesh, thus achieving effective removal of the fillet features.

\bigskip % Adds a bit of vertical space for clarity

In summary, our fillet removal method follows these key steps:
\begin{itemize}
    \item Input Model: A triangle mesh model with fillet features.
    \item Fillet Recognition: Use a pre-trained deep learning network to accurately segment all fillet faces.
    \item Boundary Mesh Extension: After removing the fillet faces, extend the resulting hole boundaries outward to form an overlapping extended mesh.
    \item Self-Intersection Resolution: Perform a Boolean intersection operation on the entire model, including the extended mesh, to resolve geometric overlaps and generate sharp edges.
    \item Mesh Cleaning: Remove non-manifold structures and isolated components resulting from the intersection to ensure the validity and topological integrity of the output model.
\end{itemize}

\begin{figure*}[htbp]
    \centering
    \includegraphics[width=0.9\textwidth]{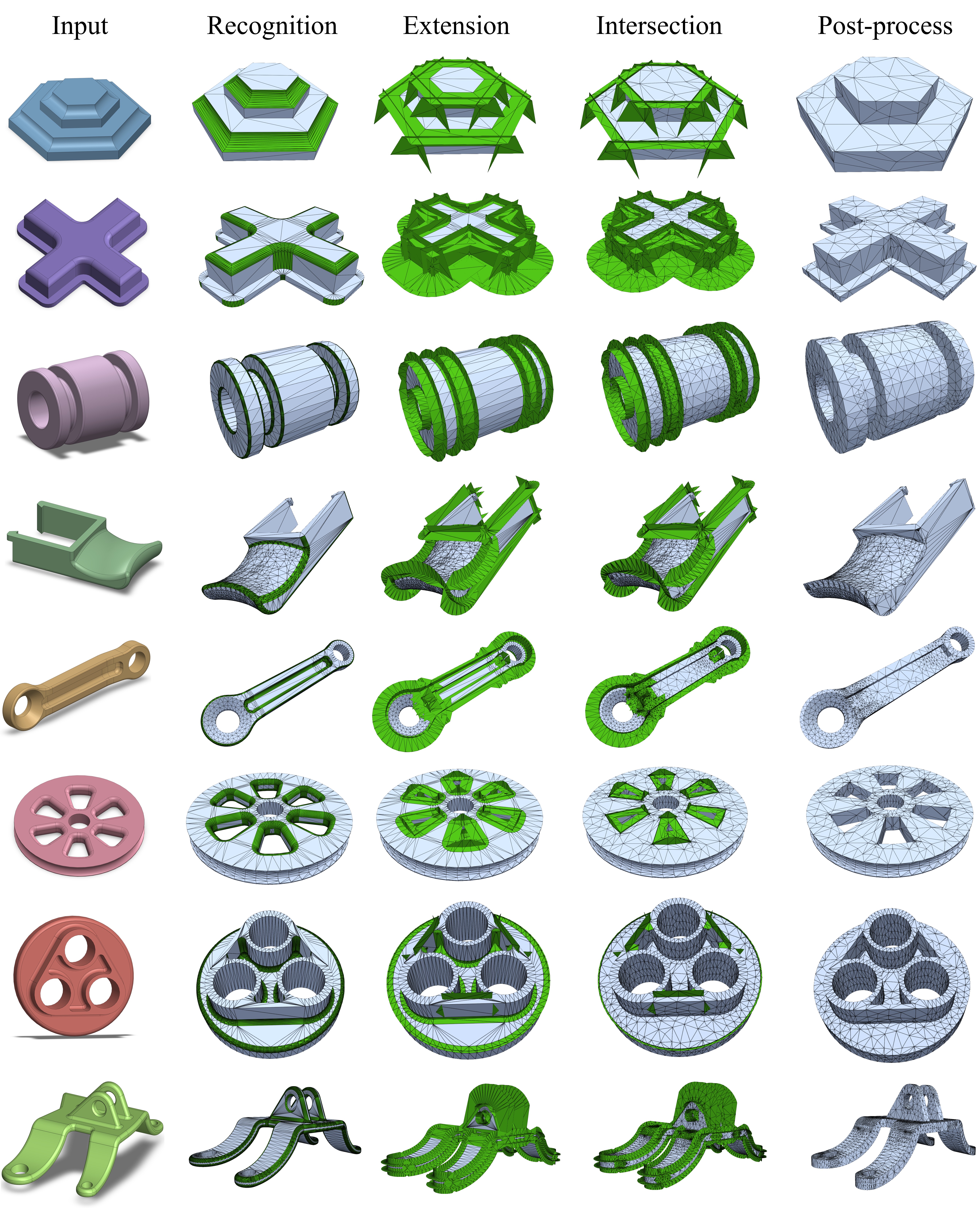} 
    \caption{Qualitative results of our end-to-end fillet removal framework on 8 challenging models from the ABC dataset. The green areas highlight the identified and processed fillet regions.}
    \label{fig:Defillet_samples}
\end{figure*}

To comprehensively evaluate the effectiveness and robustness of our end-to-end automated framework, we applied it to eight challenging models selected from the ABC dataset. As shown in Figure~\ref{fig:Defillet_samples}, these cases cover a variety of complex topological structures. In all test cases, our method successfully converted the original models with complex fillet features (leftmost model of each set) into high-quality, simplified models with clean, sharp edges (rightmost model of each set).

\subsection{Mesh Comparison}

\begin{figure}[H]
    \centering
    \includegraphics[width=0.45\textwidth]{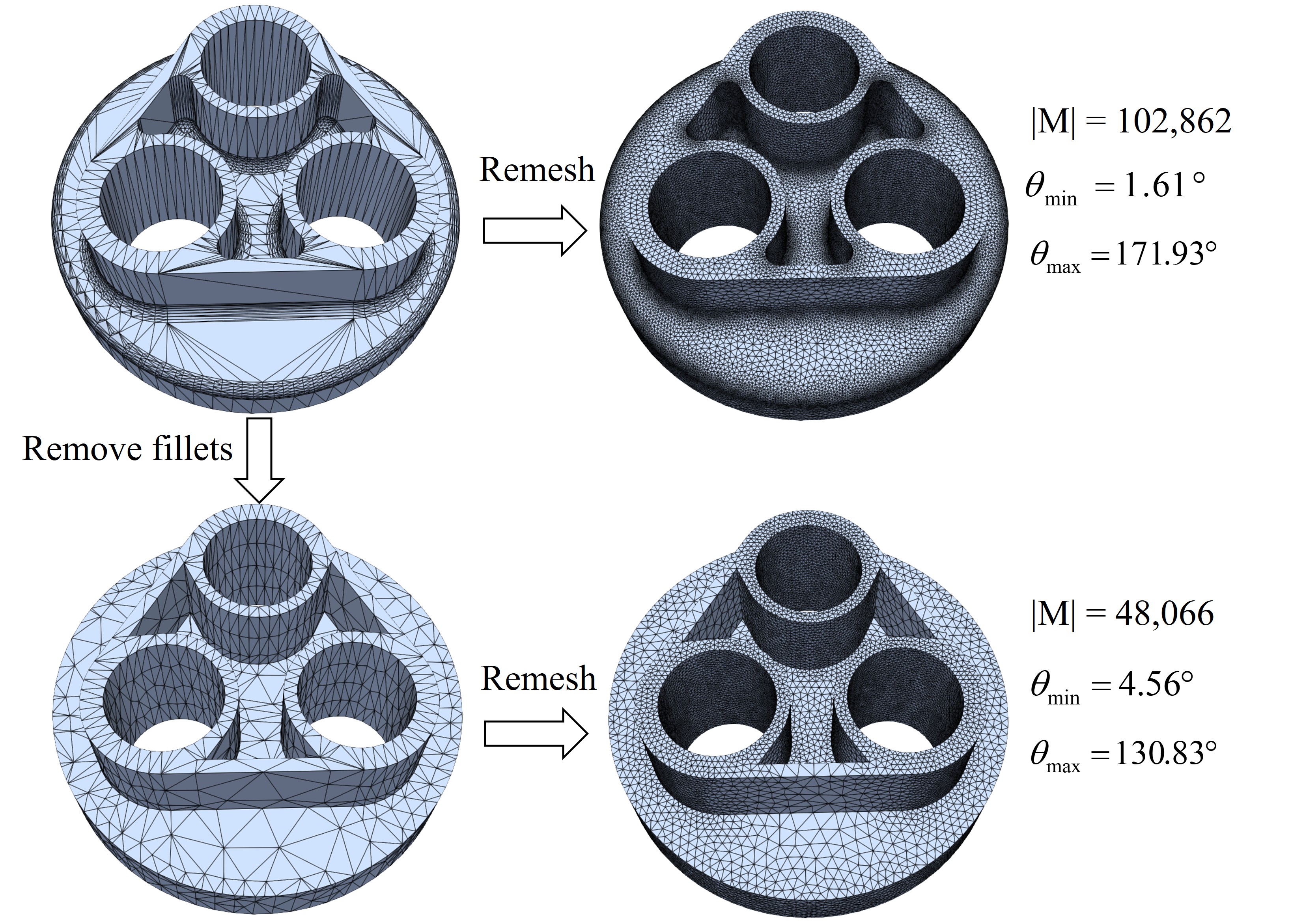} 
    \caption{Comparison of remeshing results with and without fillet removal.}
    \label{fig:remesh}
\end{figure}

To optimize the computational mesh, we employed the method by Liu et al. \cite{liuFastIntersectionfreeRemeshing2025a} to compare the remeshing outcomes on the model with and without its fillets. Both remeshing processes used the same parameters: a tolerance of $\epsilon = 10^{-4}$, a feature angle of 45$^\circ$, and 10 optimization passes. As shown in Figure \ref{fig:remesh}, removing the fillets before remeshing leads to a significant reduction in the element count (from 102,862 to 48,066) and a marked improvement in mesh quality. This approach reduces the number of mesh elements for subsequent simulations, thereby enhancing computational efficiency.

\section{Limitations}

\begin{figure}[htbp]
    \centering
    \includegraphics[width=0.3\textwidth]{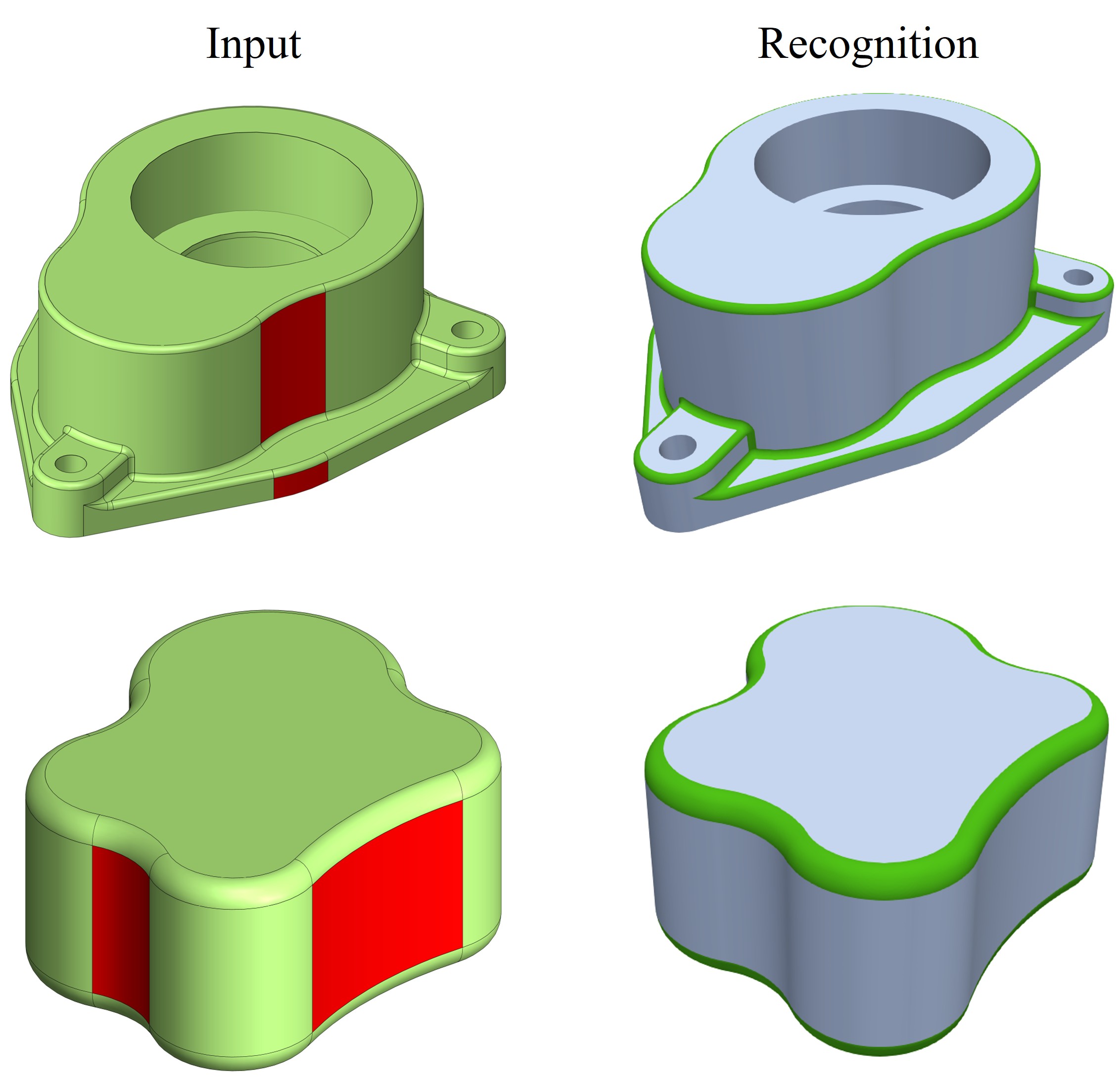} 
    \caption{Limitation of the fillet recognition network. Green indicates successfully identified fillets, while red indicates large-radius fillets that failed to be recognized.}
    \label{fig:detect_error}
\end{figure}

A current limitation of our model is its inability to effectively recognize large-radius fillets. This comes from  two main factors: first, our training dataset lacks samples of such features, leaving the model with no basis for learning to identify them. Second, our design prioritizes the handling of small-radius fillets, which have a more severe impact on mesh quality. We hypothesize that large-radius fillets, due to their gradual transitions, generally do not cause sharp increases in mesh density or produce low-quality elements. As illustrated in Figure \ref{fig:detect_error}, a typical large-radius fillet region can maintain well-shaped elements under standard meshing. Therefore, we did not include the recognition of these features within the scope of the current model.

\begin{figure}[htbp]
    \centering
    \includegraphics[width=0.3\textwidth]{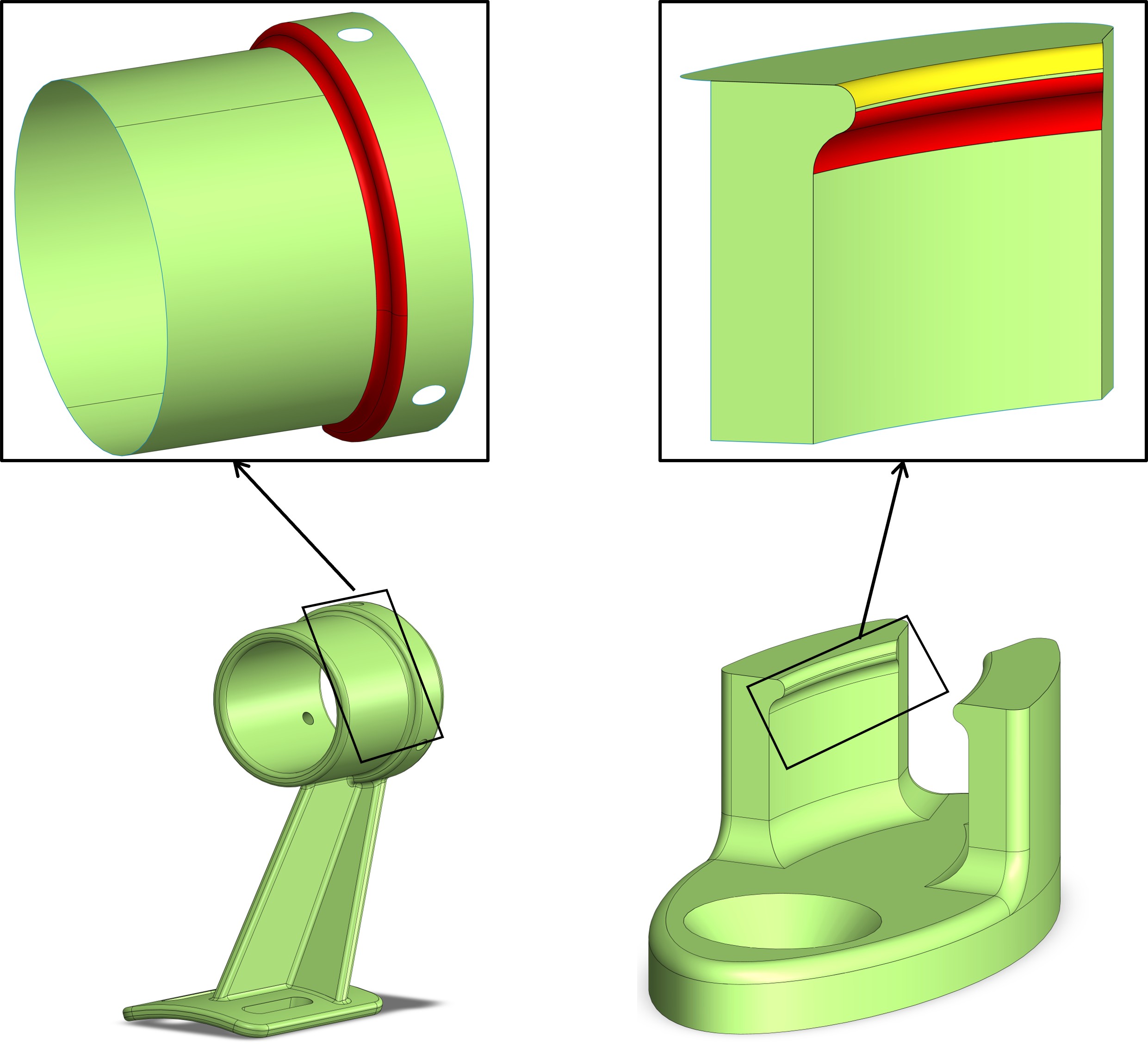} 
    \caption{Topological limitation of the fillet removal algorithm. Red indicates fillets that cannot be removed, while yellow indicates fillets that can be successfully removed.}
    \label{fig:defeature_error}
\end{figure}

Furthermore, the method fails when processing fillets (or chamfers) that connect two parallel planes, as shown in Figure \ref{fig:defeature_error}. Our core "Extend-Intersect-Clean" strategy relies on extending the adjacent surfaces to create a geometric intersection. However, when the adjacent surfaces are parallel, their extensions also remain parallel and therefore never intersect. Without this crucial intersection step, the algorithm cannot rebuild the sharp edge and thus cannot handle such topological configurations.

\section{Conclusion and Future Work}
\label{sec:conclusion}

This paper presents an integrated hybrid framework to address the long-standing challenge of automated and robust fillet recognition and simplification in CAD models. The framework combines a deep learning model for geometric understanding with a procedural geometric algorithm for precise manipulation, creating an end-to-end automated workflow from recognition to simplification.

The main contributions of this paper are three-fold. First, we built and released a large-scale, diverse benchmark dataset for fillet recognition, aiming to address the limited coverage of existing data and to provide the community with a more challenging evaluation platform. Second, we designed FilletRec, a lightweight graph neural network. By using intrinsic, rigid-transformation-invariant geometric features, FilletRec achieves high-precision recognition while showing strong generalization ability and excellent model efficiency, using far fewer parameters than existing baseline models. Finally, we combined the recognition results from FilletRec with a robust Extend-Intersect-Clean geometric algorithm, enabling effective processing of complex fillet topologies and generating simplified models suitable for downstream CAE analysis.

Experimental validation shows that our approach outperforms several state-of-the-art baseline models. It not only achieves high accuracy on our constructed dataset but also demonstrates excellent generalization ability on industrial-grade models with complex topologies. Moreover, cross-dataset testing validates the value of our new dataset for evaluating model robustness.

Looking ahead, this work can be extended in several directions. First, the model's adaptability could be improved by expanding the training data to include other types of transition features, such as chamfers. Second, to address the limitations of the current simplification algorithm in challenging cases (for example, fillets between parallel faces), future research could explore more flexible geometric reconstruction strategies. Finally, applying deep learning to the simplification step itself—for instance, using generative models to directly predict the simplified geometry is a promising direction for achieving higher levels of automation.

\bibliographystyle{elsarticle-num}
\bibliography{reference}

\end{document}